\documentclass[10pt,twocolumn,letterpaper]{article}

\usepackage{cvpr}              

\usepackage{amsmath,amsfonts,bm}

\def\eqref#1{equation~\ref{#1}}

\def\1{\bm{1}}

\DeclareMathAlphabet{\mathsfit}{\encodingdefault}{\sfdefault}{m}{sl}
\SetMathAlphabet{\mathsfit}{bold}{\encodingdefault}{\sfdefault}{bx}{n}

\usepackage{booktabs}       
\usepackage{amsfonts}       
\usepackage{nicefrac}       
\usepackage{microtype}      
\usepackage{xcolor}         

\usepackage{amsthm}
\usepackage{amsmath}
\usepackage{amssymb}
\usepackage{multirow}
\usepackage{footnote}
\usepackage{subcaption}
\usepackage{caption}
\usepackage{graphicx}
\usepackage{bbding}
\usepackage{algorithmic}
\usepackage{algorithm}
\floatname{algorithm}{Algorithm}
\usepackage{bigstrut}
\usepackage{tabularx}
\usepackage{tcolorbox}
\usepackage{colortbl}

\usepackage[table,xcdraw]{xcolor}

\usepackage{longtable}
\usepackage{array}
\usepackage{makecell}
\renewcommand{\arraystretch}{1.3}

\usepackage{listings}

\newcommand{\TODO}[1]{\textbf{\color{red}[TODO: #1]}}
\renewcommand{\TODO}[1]{}

\usepackage{fontawesome5}
\usepackage[accsupp]{axessibility} 

\definecolor{cvprblue}{rgb}{0.21,0.49,0.74}
\usepackage[breaklinks,colorlinks,allcolors=cvprblue]{hyperref}
\usepackage{titletoc}
\usepackage{multicol}

\def\paperID{13753}
\def\confName{CVPR}
\def\confYear{2026}

\title{BAMI: Training-Free Bias Mitigation in GUI Grounding}

\newcommand*\samethanks[1][\value{footnote}]{\footnotemark[#1]}

\author{
    Borui Zhang\textsuperscript{1}\thanks{Equal contribution, order decided by coin flip.} \quad
    Bo Zhang\textsuperscript{1}\samethanks \quad
    Bo Wang\textsuperscript{1}\samethanks \quad
    Wenzhao Zheng\textsuperscript{1} \quad
    Yuhao Cheng\textsuperscript{2} \\ 
    Liang Tang\textsuperscript{2} \quad
    Yiqiang Yan\textsuperscript{2} \quad
    Jie Zhou\textsuperscript{1} \quad
    Jiwen Lu\textsuperscript{1, \faEnvelope}\thanks{\faEnvelope~Corresponding author: Jiwen Lu (lujiwen@tsinghua.edu.cn).} \\[3mm]
    \textsuperscript{1}Tsinghua University, China \quad
    \textsuperscript{2}Lenovo Research, China
}

\begin{document}
\maketitle
\begin{abstract}
GUI grounding is a critical capability for enabling GUI agents to execute tasks such as clicking and dragging. 
However, in complex scenarios like the ScreenSpot-Pro benchmark, existing models often suffer from suboptimal performance. 
Utilizing the proposed \textbf{Masked Prediction Distribution (MPD)} attribution method, we identify that the primary sources of errors are twofold: 
high image resolution (leading to precision bias) and intricate interface elements (resulting in ambiguity bias). 
To address these challenges, we introduce \textbf{Bias-Aware Manipulation Inference (BAMI)}, which incorporates two key manipulations, coarse-to-fine focus and candidate selection, to effectively mitigate these biases. 
Our extensive experimental results demonstrate that BAMI significantly enhances the accuracy of various GUI grounding models in a training-free setting. 
For instance, applying our method to the TianXi-Action-7B model boosts its accuracy on the ScreenSpot-Pro benchmark from 51.9\% to 57.8\%. 
Furthermore, ablation studies confirm the robustness of the BAMI approach across diverse parameter configurations, highlighting its stability and effectiveness.
\footnote{Code is available at \url{https://github.com/Neur-IO/BAMI}.}
\end{abstract}
    
\vspace{-4mm}
\section{Introduction}

\begin{figure}[t]
    \centering
    \includegraphics[width=\linewidth]{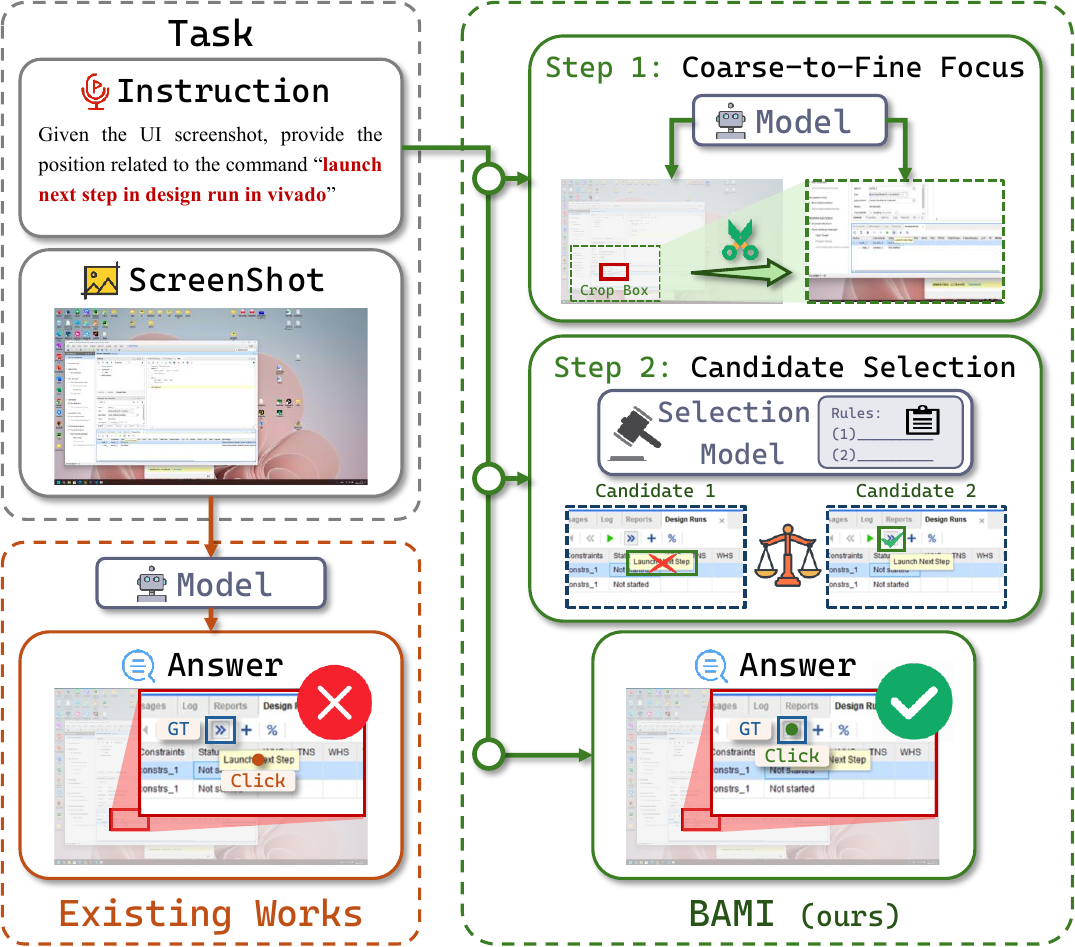}
    \vspace{-2mm}
    \caption{Compared with conventional grounding models, \textbf{BAMI} achieves accurate localization without additional training via structured inference with bias-aware manipulations.}
    \label{fig:head_a}
    \vspace{-4mm}
\end{figure}

The advent of multimodal large language models (MLLMs)~\cite{hurst2024gpt,bai2025qwen2} has made it increasingly feasible for GUI agents to automate tasks across desktop and mobile platforms. 
At the core of these agents lies \emph{GUI Grounding}:
given a pair of \emph{natural language instructions} and a \emph{screenshot}, the task is to accurately localize the coordinates of the target element within a high-resolution graphical interface, 
thereby enabling subsequent atomic actions such as clicking, typing, or dragging. 
Early approaches often relied on structured interface representations, such as XML or DOM trees~\cite{deng2023mind2web,gurreal}. 
However, these structures are frequently unavailable or inconsistent with the visual rendering in real-world scenarios. 
Consequently, research has shifted toward the visual paradigm of \emph{instruction + screenshot}, where MLLMs directly output coordinates~\cite{wu2024atlas,xu2024aguvis,lu2024omniparser,gounavigating2025,qin2025ui}, providing a more robust perceptual foundation for agents.
In comparison to general natural image tasks, GUI scenarios present unique challenges due to their \textbf{high resolution} and \textbf{dense elements}, where semantics are determined by a combination of icons, text, and contextual cues. 
These characteristics make accurate localization significantly more challenging. 
For instance, in ScreenSpot-Pro~\cite{li2025screenspot}, a benchmark dataset covering professional software across multiple domains, the localization accuracy of most models remains below 50\%.

The performance of multimodal grounding models remains underutilized. 
In particular, performance improvements can be achieved \textbf{without additional training} by optimizing inference methods. 
From an error-driven perspective, we categorize grounding failures into two primary types:
\textbf{(1) Knowledge deficiency}: The model fails to recognize the target due to a lack of relevant knowledge.
\textbf{(2) Inductive bias}: The model has the necessary knowledge but makes errors due to its inherent selection bias, which manifests in two typical forms, namely \textit{precision bias} and \textit{ambiguity bias}.
To diagnose these causes of failure, we introduce a \textbf{Masked Prediction Distribution (MPD)} method. 
This approach randomly occludes parts of the screenshot, makes repeated predictions, and aggregates the frequency of hotspots or candidate points across the image. 
This aggregation reveals how the model distributes its focus across the image. Statistical analysis of 50 error samples shows that approximately 14\% of failures stem from knowledge deficiency, while 74\% are attributed to inductive bias.

\begin{figure}[t]
    \centering
    \includegraphics[width=\linewidth]{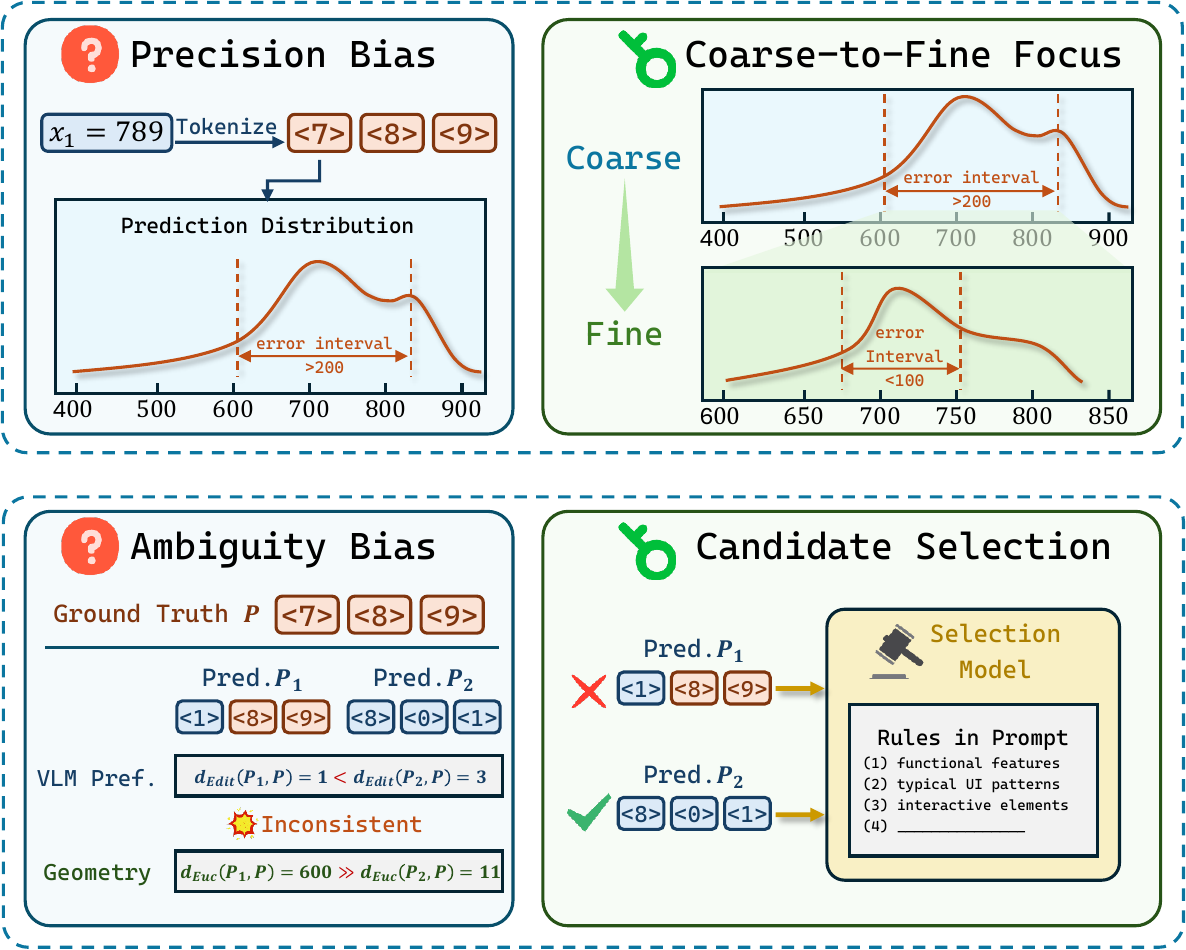}
    \vspace{-2mm}
    \caption{\textbf{Bias Mitigation Strategy.} To address accuracy bias and ambiguity bias, BAMI introduces two manipulations: coarse-to-fine focus and candidate selection.}
    \label{fig:head_b}
\end{figure}

\begin{figure}[t]
    \centering
    \includegraphics[width=0.65\linewidth]{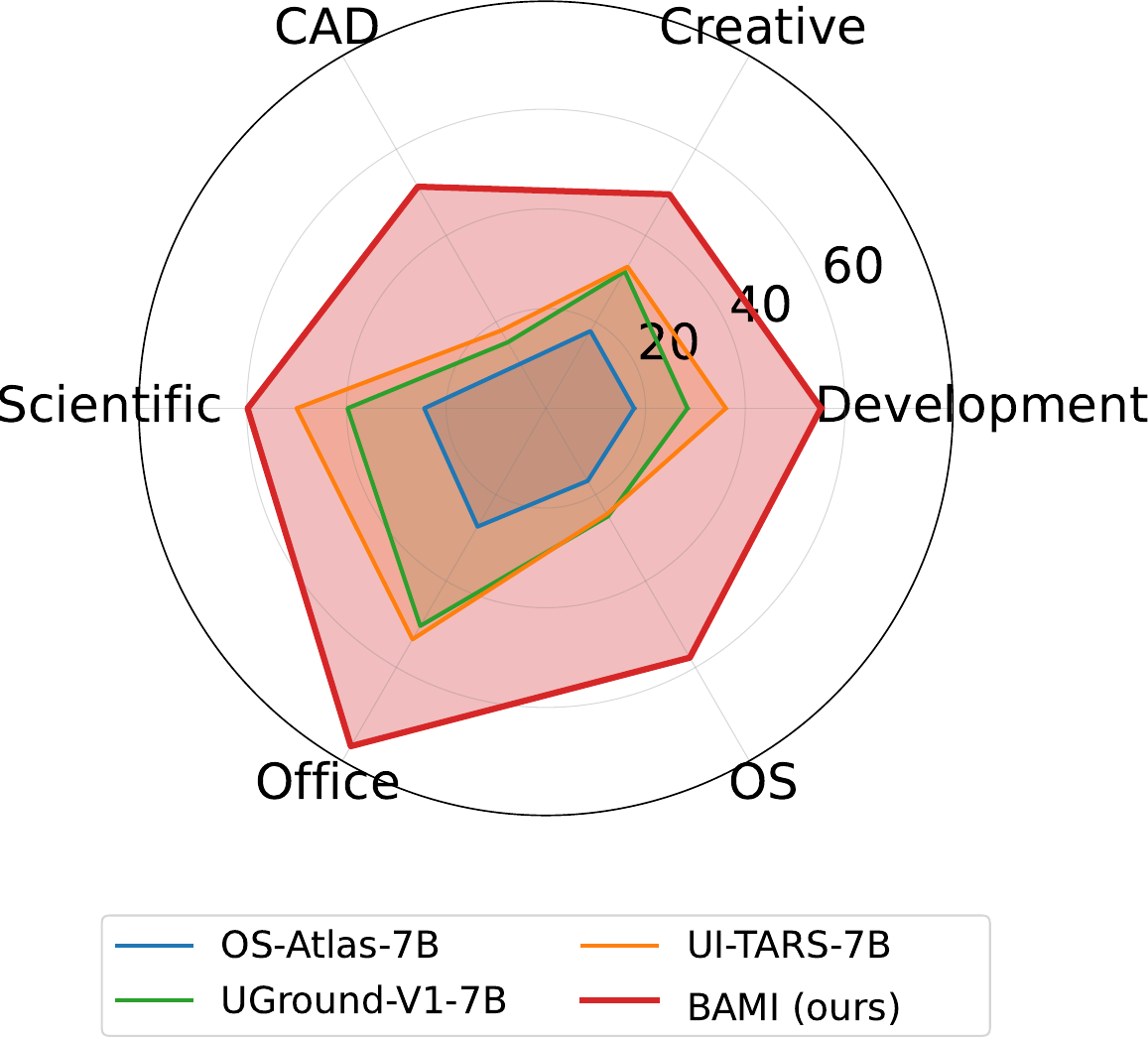}
    \vspace{-2mm}
    \caption{\textbf{Accuracy comparison on ScreenSpot-Pro.} BAMI consistently improves performance across all model backbones.}
    \label{fig:radar_plot}
    \vspace{-3mm}
\end{figure}

In this paper, we propose \textbf{Bias-Aware Manipulation Inference (BAMI)}. 
The key idea is to transform the one-step localization task into a recursive, multi-step structured inference process through predefined bias-aware manipulations (Figure~\ref{fig:head_a} and~\ref{fig:head_b}). 
To mitigate precision bias, we decompose localization into hierarchical \textbf{coarse-to-fine focus}, 
where each step refines the candidate region identified in the previous round.
This progressive refinement reduces the search space and improves the resolution of the predicted coordinates. 
To address ambiguity bias, we incorporate an external \textbf{Candidate Selection}. 
By defining selection rules specific to the localization task and injecting these rules into the model as prompts, 
we correct the model's erroneous selection preferences. 
Importantly, our method does not require any additional model training and can be directly applied to a variety of existing open-source backbones.
We evaluate BAMI on multiple open-source backbones (e.g., OS-Atlas-7B~\cite{wu2024atlas}, UI-TARS-7B~\cite{qin2025ui}, and TianXi-Action-7B~\cite{tang2025sea}) and several datasets (e.g., ScreenSpot-Pro~\cite{li2025screenspot}, ScreenSpot-V2~\cite{wu2024atlas}). 
BAMI consistently improves accuracy on complex samples (Figure~\ref{fig:radar_plot}). 
Ablation studies further confirm the effects of \textbf{coarse-to-fine focus} and \textbf{candidate selection}. 
Our results demonstrate that extending and structuring the reasoning path during inference provides a cost-effective means of unlocking the full grounding potential of existing models.
The main contributions of this work are as follows:
\begin{itemize}
    \item \textbf{Diagnosis of Grounding Failures}: We introduce the MPD method to diagnose common grounding failures, such as knowledge deficiency and inductive bias.
    \item \textbf{Precision Bias Mitigation}: We transform single-step localization into a multi-step progressive search through hierarchical cropping, which effectively reduces precision bias in high-resolution and small-object scenarios.
    \item \textbf{Ambiguity Bias Correction}: To address discrepancies between MLLM's edit distance and spatial coordinate distance, we introduce an external selection and correct the MLLM's selection bias using predefined prompt rules.
    \item \textbf{Training-free Improvements}: We validate BAMI across various backbones and benchmarks, demonstrating consistent improvements and emphasizing the general value of test-time reasoning design in GUI Grounding.
\end{itemize}

\section{Related Work}
Training on pre-trained MLLMs~\cite{bai2025qwen2} has been demonstrated to significantly enhance GUI grounding capabilities. 
Early approaches predominantly relied on conventional instruction fine-tuning.
With the introduction of DeepSeek-R1~\cite{guo2025deepseek}, reinforcement learning fine-tuning has attracted growing attention. 
Meanwhile, several studies have found that specially designed inference methods help tap into the potential of MLLMs in terms of localization capabilities.
\subsection{Instruction Fine-tuning}
The simplest approach is to fine-tune pre-trained MLLMs (e.g., Qwen2.5-VL~\cite{bai2025qwen2}) on task-specific GUI instruction datasets. 
Early work such as AGUVIS~\cite{xu2024aguvis} introduced vision-based models for GUI grounding.
To address high-resolution GUI screenshots, CogAgent~\cite{hong2024cogagent} introduced a cross-resolution efficient attention mechanism.
ShowUI~\cite{lin2025showui} applied token pruning based on GUI interface structure, improving both efficiency and performance. 
OmniParser~\cite{lu2024omniparser} converted GUI pixels into structured tokens that could be parsed by LLMs.
In terms of dataset construction, SeeClick~\cite{cheng2024seeclick} proposed an automated pipeline for managing GUI data.
UGround~\cite{gounavigating2025} built a large-scale dataset with 10M elements, improving generalization. 
With the advent of larger-scale datasets and more powerful models, new large-scale systems such as UI-TARS~\cite{qin2025ui} and Phi-Ground~\cite{zhang2025phi} have pushed the SOTA performance across benchmarks.

\subsection{Reinforcement Learning}
Given the fine-grained nature of GUI localization, instruction fine-tuning alone is often insufficient for achieving high precision. 
DeepSeek-R1~\cite{guo2025deepseek} introduced the GRPO method, demonstrating the potential of reinforcement learning in enhancing spatial reasoning for GUI grounding tasks. 
Following this, UI-R1~\cite{lu2025ui} and GUI-R1~\cite{luo2025gui} were among the first to apply GRPO in GUI tasks. 
InfiGUI-R1~\cite{liu2025infigui} focused on reward function design, emphasizing IoU-based metrics to improve localization accuracy. 
GUI-G1~\cite{zhou2025gui} introduced box-attribute constraints to regulate bounding-box geometry, 
while GUI-G2~\cite{tang2025gui} modeled spatial distributions using Gaussian functions. 
TianXi-Action~\cite{tang2025sea} focused on generating high-quality reinforcement learning data.
Collectively, these studies affirm the efficacy of reinforcement learning in enhancing spatial reasoning in GUI tasks.

\subsection{Inference Enhancement}
Significant attention has been given to optimizing inference strategies to exploit the capabilities of MLLMs. 
One line of work extends reasoning chains in the language space; however, experiments~\cite{zhang2025does} have found this direction suboptimal for GUI scenarios, 
sometimes even hindering performance. 
Alternatively, several works have targeted inference enhancement in the image space. 
ScreenSeekeR~\cite{li2025screenspot} and R-VLM~\cite{park2025r} introduced multi-stage pipelines, 
first performing region-level localization followed by refinement within local regions, thus improving accuracy. 
DiMo-GUI~\cite{wu2025dimo} proposed a divide-and-conquer strategy, separating reasoning over icons and text to reduce cross-modal interference. 
GUI-RC~\cite{du2025test} employed intersection operations to aggregate multiple predictions, improving robustness. 
While conventional MLLMs have demonstrated the effectiveness of inference enhancement techniques for general tasks~\cite{liu2024paying}, their direct application to GUI tasks is often limited by inductive biases specific to spatial reasoning. 
This paper identifies two critical inductive biases
—\textbf{precision bias and ambiguity bias}—
that remain prominent in GUI grounding. 
We propose BAMI to address these through bias-aware inference.
\section{Pilot Study}
On ScreenSpot-Pro~\cite{li2025screenspot}, a challenging GUI grounding benchmarks, the accuracy of state-of-the-art grounding models on these benchmarks has significantly decreased, falling below 50\%. 
To gain deeper insights into the underlying performance bottlenecks, we conducted a systematic pilot study addressing two primary questions: 
\emph{(1) What are the root causes of errors made by GUI grounding models?
(2) How can these errors be mitigated from a model mechanism perspective without the need for retraining?}
\subsection{Error Attribution via MPD}
This section uses the ScreenSpot-Pro dataset~\cite{li2025screenspot} as a benchmark to analyze potential error patterns in GUI grounding models and explore corresponding mitigation strategies.
\begin{figure*}[t]
    \centering
    \begin{subfigure}{0.25\textwidth}
        \centering
        \includegraphics[width=\textwidth]{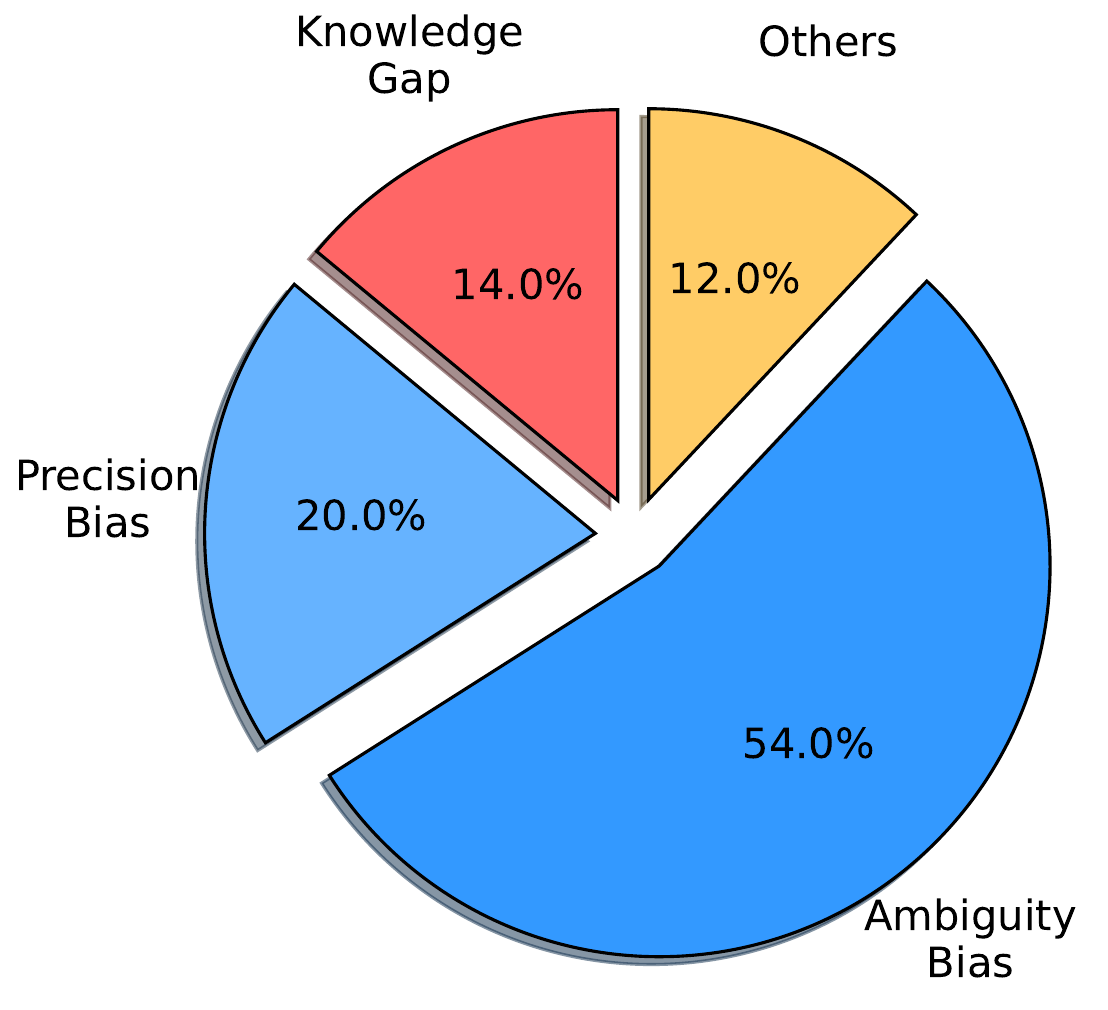}
        \caption{}
        \label{fig:pilot_a}
    \end{subfigure}
    \hfill
    \begin{subfigure}{0.74\textwidth}
        \centering
        \includegraphics[width=\textwidth]{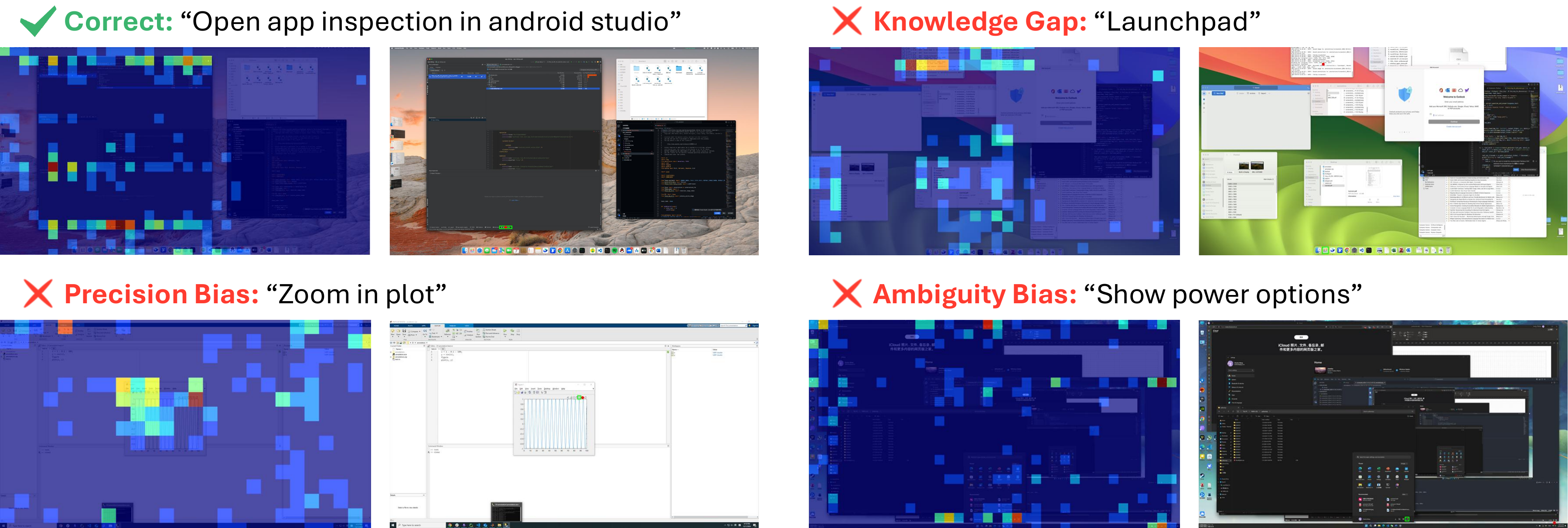}
        \caption{}
        \label{fig:pilot_b}
    \end{subfigure}
    \vspace{-2mm}
    \caption{\textbf{Error Attribution Analysis.} (a) Proportions of attribution types. (b) Attribution analysis of model predictions. The deep red regions in the heatmap indicate potential prediction locations, demonstrating how the MPD can clearly identify the sources of model errors.}
    \label{fig:pilot_ab}
    \vspace{-4mm}
\end{figure*}

\paragraph{Problem Formulation}
For a GUI grounding model $f$, given a query $q$ and a GUI screenshot $I \in \mathbb{R}^{H \times W \times 3}$, the model generates a text sequence $t$ containing the target bounding box in the standard format:
\texttt{<|box\_start|>$(x_1, y_1, x_2, y_2)$<|box\_end|>}. 
The coordinates $(x_1, y_1, x_2, y_2) = r(t)$ are extracted using a regular expression parser $r$, where $(x_1, y_1)$ and $(x_2, y_2)$ represent the top-left and bottom-right coordinates of the bounding box, respectively. The center coordinates of the bounding box are computed as:
$(x_c, y_c) = \left(\frac{x_1 + x_2}{2}, \frac{y_1 + y_2}{2}\right)$.
A prediction is considered correct if the center coordinate $(x_c, y_c)$ lies within the ground-truth bounding box; otherwise, it is deemed an error.
\paragraph{Attribution Method}
Traditional gradient-based attribution methods (e.g., GradCAM~\cite{selvaraju2017grad}, Integrated Gradients~\cite{sundararajan2017axiomatic}) are not well-suited for the discrete text-to-coordinate conversion process. 
As an alternative, we initially considered using Shapley values~\cite{shapley1953value, lundberg2017unified} for attribution analysis. 
For an $n$-dimensional input feature, the Shapley value for the $i$-th feature is defined as:
$\phi_i = \sum_{S \subseteq \{1,2,\ldots,n\} \setminus \{i\}} \frac{|S|!(n - |S| - 1)!}{n!} \left[f(S \cup \{i\}) - f(S)\right]$,
where $S$ denotes a subset of features. 
However, due to the high resolution of GUI screenshots, estimating the Shapley values~\cite{ancona2019explaining} for a single sample takes approximately 10 hours on a single RTX 4090 GPU, which is computationally impractical. 
To address this, we propose the \textbf{Masked Prediction Distribution (MPD)} method, which efficiently observes the spatial distribution of model predictions under random perturbations (see Supplementary Algorithm 1). 
Regions with densely distributed predicted points indicate high model confidence in those areas.
We set the number of perturbations to 300 per sample and can obtain heatmaps within 20 minutes per sample.
\paragraph{Error Pattern Analysis}
Based on the experimental results of TianXi-Action-7B~\cite{tang2025sea} on ScreenSpot-Pro, 
we conducted an attribution analysis on 50 error samples, with the findings summarized in Table~\ref{tab:error-analysis}. 
Notably, both precision bias and ambiguity bias are categorized as inductive bias issues, collectively accounting for 74\% of the error samples. 
This indicates that if we can effectively mitigate inductive bias, the model's performance will be significantly improved.
\subsection{Mitigation Strategy: Inductive Bias Correction}
Based on the error pattern analysis, we explored potential mitigation methods for different error types. 
Knowledge gap errors reflect limitations in the model's training data or architecture, which are difficult to address with inference-time techniques. 
In contrast, inductive bias errors (precision bias and ambiguity bias) can potentially be mitigated through optimization of the inference mechanism.
\begin{table}[t]
    \centering
    \small
    \setlength{\abovecaptionskip}{4pt}
    \setlength{\belowcaptionskip}{4pt}
    \caption{Proportions and detailed analysis of different error types.}
    \vspace{-2mm}
    \label{tab:error-analysis}
    \setlength{\arrayrulewidth}{0.5pt}
    \setlength{\belowrulesep}{0pt}
    \setlength{\aboverulesep}{0pt}
    \renewcommand{\arraystretch}{1.25}
    \begin{tabular}{>{\centering\arraybackslash}m{2.5cm}m{5.5cm}}
        \rowcolor[HTML]{2D5F91}
        \multicolumn{1}{c}{\textcolor{white}{\textbf{Error Type}}} & 
        \textcolor{white}{\textbf{Description}} \\ \midrule
        \rowcolor[HTML]{F0F8FF}
        \makecell[c]{Knowledge Gap \\ (14\%)} & 
        Model fails to recognize target information. 7 error samples. \\
        \rowcolor[HTML]{FFFFFF}
        \makecell[c]{Precision Bias \\ (20\%)} & 
        Model identifies target but exhibits systematic offset. 10 samples. \\
        \rowcolor[HTML]{F0F8FF}
        \makecell[c]{Ambiguity Bias \\ (54\%)} & 
        Model distracted by similar regions or misleading semantics. 27 samples. \\
        \rowcolor[HTML]{FFFFFF}
        \makecell[c]{Others (12\%)} & 
        Unclassified patterns. 6 samples. \\
        \bottomrule
    \end{tabular}
    \vspace{-3mm}
\end{table}
\paragraph{Limitations of Language-Space Enhancement}
Inspired by reasoning techniques in large language models (e.g., Chain-of-Thought~\cite{wei2022chain}), we first attempted to enhance GUI grounding performance by augmenting linguistic information. 
\textbf{(1) Query Expansion Strategy:} For queries with insufficient or ambiguous descriptions, we used a language model to expand and refine the original query, generating more precise instruction information.
\textbf{(2) Context Expansion Strategy:} We utilized a multimodal large language model (e.g., Qwen2.5-VL~\cite{bai2025qwen2}) to generate a structured description of the GUI, including the geometric location, text content, and other information of UI elements, and concatenated this with the original query as model input.
However, experimental results indicated that merely extending the language sequence did not significantly improve model accuracy, and even introduced additional errors in some cases. 
This phenomenon aligns with recent findings~\cite{zhang2025does} that traditional linguistic reasoning models are difficult to directly transfer to precise grounding tasks.
\paragraph{Root Causes of Precision Bias}
An in-depth analysis of precision bias revealed that multimodal models typically adopt discretized coordinate representations for images with resolution $H \times W$. 
For instance, in Qwen series models, a coordinate value of $x_1 = 789$ is split into independent digit characters (\texttt{<7>}, \texttt{<8>}, \texttt{<9>}) and further converted into their corresponding token IDs. 
This discretization inherently limits the model's maximum precision to the unit digit level.
\paragraph{Root Causes of Ambiguity Bias}
The cross-entropy training objective for multimodal models optimizes the \textbf{edit distance} of token sequences rather than the \textbf{Euclidean distance}. 
Let the ground-truth coordinate be $x_{\text{GT}} = 789$, and consider two predicted candidates: $x' = 189$ and $x'' = 801$. 
A direct comparison of the two metrics yields:
\begin{align*}
    d_{\text{edit}}(x_{\text{GT}}, x') &= 1 < d_{\text{edit}}(x_{\text{GT}}, x'') = 3 \\
    d_{\text{euc}}(x_{\text{GT}}, x') &= 600 > d_{\text{euc}}(x_{\text{GT}}, x'') = 12
\end{align*}
This inconsistency in metrics causes a fundamental conflict between the model's optimization objective in token space and the need for accuracy in real-world spatial localization. Therefore, external correction mechanisms combining token sequence optimization with geometric constraints are necessary to address this systematic bias.

\section{Method}
Based on the experimental results from the pilot study, we design the \textbf{BAMI} method in this section. 
The method targets both accuracy bias and ambiguity bias, and proposes different manipulations to improve GUI grounding.
\subsection{Accuracy Bias: Coarse-to-Fine Focus}
The root cause of accuracy bias lies in the discretization process of multimodal large models during coordinate localization. 
Since the prediction accuracy of the model is typically limited to the pixel level, and its output is difficult to be perfectly accurate, prediction errors may sometimes reach tens or even hundreds of pixels. 
To effectively eliminate accuracy bias, inspired by human observation strategies, we propose a \textbf{coarse-to-fine focus} manipulation.
Specifically, we first use the grounding model to predict a coarse localization coordinate $(x^t, y^t)$. 
Then, based on this coarse coordinate, we crop the original image to a scale of $\lambda < 1$, 
and input the cropped image back into the grounding model for fine localization, obtaining a more precise coordinate $(x^{t+1}, y^{t+1})$. 
Although this process can be iterated multiple times, we find that there is a trade-off in the hyperparameters.  
\textbf{(1) Iteration count:} After a certain number of iterations, the performance improvement of the model tends to plateau;  
\textbf{(2) Crop ratio:} A large cropping ratio may lead to the loss of crucial information, while a small cropping ratio may prevent the model from accurately localizing the target. 
\begin{figure*}[t]
    \centering
    \includegraphics[width=\textwidth]{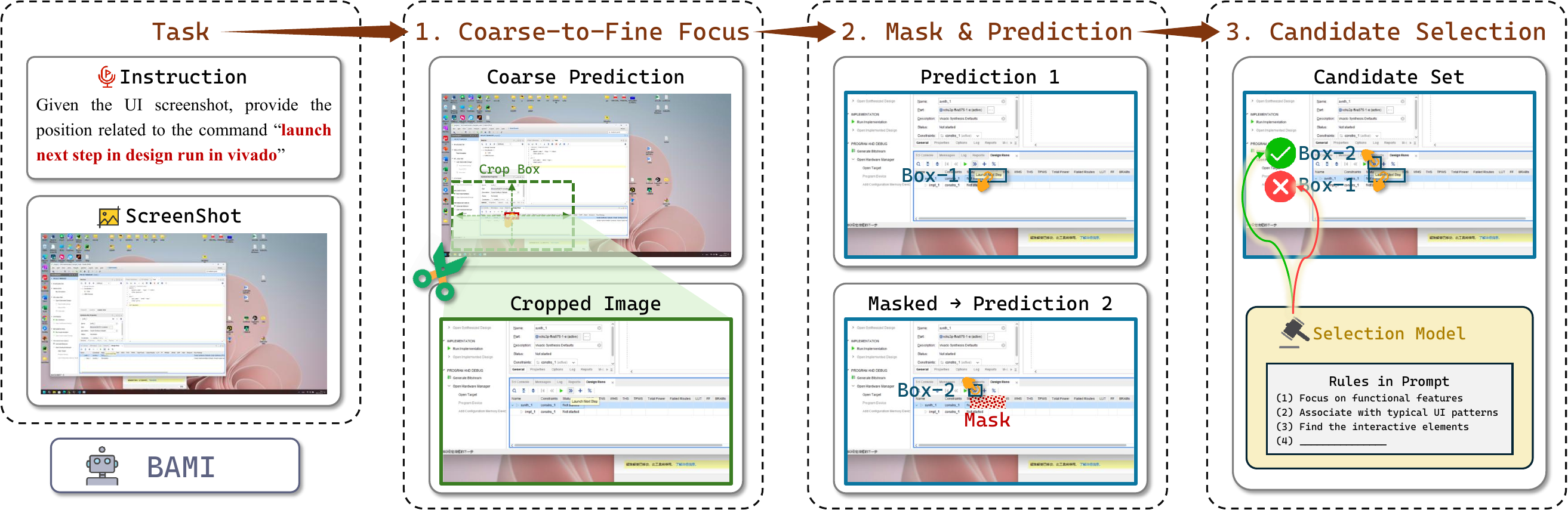}
    \vspace{-2mm}
    \caption{\textbf{Illustration of BAMI.} 
    \textbf{Step 1:} Based on the initial prediction results of the grounding model, BAMI performs cropping around these initial predictions at a predefined ratio. 
    \textbf{Step 2:} The model conducts multiple predictions on the cropped images; after each prediction, the pixels within the predicted bounding box are randomly masked to ensure the diversity of multiple prediction results. 
    \textbf{Step 3:} Using predefined rules and an external knowledge model, the model ranks multiple candidate coordinates and selects the final coordinates.}
    \label{fig:method_bami}
\end{figure*}

\begin{algorithm}[t]
    \caption{BAMI (with $N$ crop iterations and $M$ candidates per iteration)}
    \label{alg:bami}
    \begin{algorithmic}[1]
        \REQUIRE Query $q$, screenshot $I$, correction model $m$, and grounding model $f$
        \ENSURE Grounding point $(x, y)$
        \STATE Initialize the input image as $I^1 = I$
        \FORALL{$t \in \{1, 2, \cdots, N\}$}
            \STATE Initialize the candidate box set $\Phi^t = \emptyset$
            \FORALL{$i \in \{1, 2, \cdots, M\}$}
                \STATE Masking all pixels in the candidate set to get input image $I^t_i = \text{MASK}(I^{t-1}, \Phi^t)$ 
                \STATE Predict the candidate box $b^t_i = f(q, I^t_i)$ and update $\Phi^t \leftarrow \Phi^t \cup \{b^t_i\}$
            \ENDFOR
            \STATE Select the preferred box $\tilde{b}^t = m(q, I^t, \Phi^t)$
            \STATE Crop the input image $I^{t+1} = \text{CROP}(I^t, \tilde{b}^t)$
        \ENDFOR
        \STATE Compute the center point of $\tilde{b}^N$ as $(x, y)$
    \end{algorithmic}
\end{algorithm}
\subsection{Ambiguity Bias: Candidate Correction}
Multimodal large models (MLLMs) represent coordinates as text sequences for autoregressive generation. 
While this design simplifies the training process, it introduces a discrepancy between the training and inference phases. 
For example, the coordinate ``789'' is encoded into the text sequence \texttt{<7><8><9>}, and the model minimizes the edit distance of this text sequence using cross-entropy loss. 
In practice, however, the impact of digit position errors is asymmetric: 
an error in the hundreds place is two orders of magnitude more significant than that in the ones place. 
This results in a substantial mismatch between edit distance and Euclidean distance, and no straightforward mapping exists to convert the former to the latter.
To eliminate ambiguity bias, we first generate multiple mutually exclusive candidate bounding boxes through multi-round masked prediction operations. 
Subsequently, we utilize a correction model to re-select from these candidate boxes. 
We investigate both online APIs (e.g. GPT-5) and locally trained models (e.g. Qwen3-VL-8B) for this role.
Notably, the key to this operation lies in prompt design; naive prompt designs fail to leverage the correction model effectively. 
To enable the correction model to rectify the erroneous ordering tendency of the grounding model, 
we incorporate \textbf{key principles} consistent with GUI priors into the prompt. 
Examples of these principles are provided below, and detailed prompt design is available in the appendix.
\begin{lstlisting}[title=Prompt]
- (Functional Preference) Focus on the functional purpose of the highlighted elements
- (Memory Comparison) Consider standard patterns (e.g., buttons for actions)
- (Interactive Components) Prioritize interactive elements over static text/labels
\end{lstlisting}
\subsection{BAMI: Bias-Aware Manipulation Inference}
By integrating the two manipulations outlined above, we propose \textbf{BAMI}, as illustrated in Figure~\ref{fig:method_bami}. 
To enhance the diversity of candidate boxes, we mask the pixels within already predicted candidate boxes prior to each new prediction step, thereby ensuring the mutual exclusivity between newly generated candidate boxes and existing results. 
To mitigate precision bias, BAMI adopts a coarse-to-fine focus strategy in its outer loop, enabling gradual refinement of focus toward more accurate coordinate positions step by step. 
Simultaneously, to address ambiguity bias, BAMI employs a candidate selection strategy in each iteration, selecting the most suitable box from multiple candidates as the final output. 
The algorithm of BAMI is detailed in Algorithm~\ref{alg:bami}.

\section{Experiment}
\subsection{Experimental Setup}
\paragraph{Models} 
The proposed BAMI method aims to enhance the accuracy of Grounding models without retraining. 
We tested this method on several state-of-the-art grounding models, including OS-Atlas-7B~\cite{wu2024atlas}, UI-TARS-1.5-7B~\cite{qin2025ui}, and TianXi-Action-7B~\cite{tang2025sea}. 
All models were implemented using the Transformers framework~\cite{wolf2019huggingface} for inference. 
The input to the models consists of both the query and the screenshot. 
OS-Atlas and TianXi-Action output bounding box coordinates, while UI-TARS outputs click coordinates.
\paragraph{Data} 
We evaluate BAMI on ScreenSpot-V2~\cite{wu2024atlas}, and ScreenSpot-Pro~\cite{li2025screenspot}. 
ScreenSpot-V2 are mainly used to assess grounding accuracy in simple scenarios, covering mobile, web, and desktop. 
ScreenSpot-Pro focuses on complex scenarios, consisting of high-resolution screenshots of professional software, where each sample contains multiple software elements, and the targets are typically small, making it a particularly challenging task.
\paragraph{Hyperparameters} 
To balance efficiency and accuracy, 
two iterations were adopted for the coarse-to-fine focusing process. 
For high-resolution screenshots, the crop ratio $\lambda$ was set to the range $[0.5, 0.7]$. 
To eliminate ambiguity bias, a masking mechanism was employed, which generates $2\sim 3$ candidate results per iteration; 
subsequently, a correction model was used to select the result most relevant to the query. 
We evaluate both online (GPT-5) and offline (Qwen3-VL-8B) variants.
All experiments were conducted on a single RTX 4090 GPU.
\begin{table*}[t]
    \small
    \setlength{\tabcolsep}{3.5pt}
    \renewcommand{\arraystretch}{0.88}
    \centering
    \caption{Comparison with various models on ScreenSpot-Pro.}
    \vspace{-2mm}
        \begin{tabular}{l|cc|cc|cc|cc|cc|cc|c}
        \toprule
        \multicolumn{1}{c|}{\multirow{2}[4]{*}{\textbf{Grounding Model}}} & \multicolumn{2}{c|}{\textcolor[rgb]{ .2,  .2,  .2}{\textbf{Development }}} & \multicolumn{2}{c|}{\textcolor[rgb]{ .2,  .2,  .2}{\textbf{Creative}}} & \multicolumn{2}{c|}{\textbf{CAD}} & \multicolumn{2}{c|}{\textcolor[rgb]{ .2,  .2,  .2}{\textbf{Scientific}}} & \multicolumn{2}{c|}{\textcolor[rgb]{ .2,  .2,  .2}{\textbf{Office}}} & \multicolumn{2}{c|}{\textcolor[rgb]{ .2,  .2,  .2}{\textbf{OS}}} & \multirow{2}[4]{*}{\textbf{Avg.}} \\
    \cmidrule{2-13}          & \textbf{Text} & \textbf{Icon} & \textbf{Text} & \textbf{Icon} & \textbf{Text} & \textbf{Icon} & \textbf{Text} & \textbf{Icon} & \textbf{Text} & \textbf{Icon} & \textbf{Text} & \textbf{Icon} &  \\
        \midrule
        \multicolumn{14}{c}{\textbf{Proprietary Models}} \\
        \midrule
        GPT-4o~\cite{hurst2024gpt} & 2.0   & 0.0   & 1.3   & 0.0   & 1.0   & 0.0   & 2.1   & 0.0   & 1.1   & 0.0   & 0.0   & 0.0   & 0.8  \\
        Claude Computer Use~\cite{hu2024dawn} & 14.5  & 3.7   & 22.0  & 3.9   & 25.9  & 3.4   & 33.9  & 15.8  & 30.1  & 16.3  & 11.0  & 4.5   & 17.1  \\
        \midrule
        \multicolumn{14}{c}{\textbf{General Open-source Models}} \\
        \midrule
        Qwen2.5-VL-3B~\cite{bai2025qwen2} & 9.1   & 7.3   & 22.1  & 1.4   & 26.8  & 2.1   & 38.2  & 7.3   & 33.9  & 15.1  & 10.3  & 1.1   & 16.1  \\
        Qwen2.5-VL-7B~\cite{bai2025qwen2} & 16.8  & 1.6   & 46.8  & 4.1   & 35.9  & 7.7   & 49.3  & 7.3   & 52.5  & 20.8  & 37.4  & 6.7   & 26.8  \\
        \midrule
        \multicolumn{14}{c}{\textbf{GUI-specific Models (SFT)}} \\
        \midrule
        SeeClick-9.6B~\cite{cheng2024seeclick} & 2.5   & 0.0   & 0.6   & 0.0   & 1.0   & 0.0   & 3.5   & 0.0   & 1.1   & 0.0   & 2.8   & 0.0   & 1.1  \\
        CogAgent-18B~\cite{hong2024cogagent} & 7.1   & 3.1   & 14.9  & 0.7   & 9.6   & 0.0   & 22.2  & 1.8   & 13.0  & 0.0   & 5.6   & 0.0   & 7.7  \\
        OS-Atlas-7B~\cite{wu2024atlas} & 12.2  & 4.7   & 33.1  & 1.4   & 28.8  & 2.8   & 37.5  & 7.3   & 33.9  & 5.7   & 27.1  & 4.5   & 18.9  \\
        ShowUI-2B~\cite{lin2025showui} & 2.5   & 0.0   & 16.9  & 1.4   & 9.1   & 0.0   & 13.2  & 7.3   & 15.3  & 7.5   & 10.3  & 2.2   & 7.7  \\
        UGround-7B~\cite{gounavigating2025} & 14.2  & 1.6   & 26.6  & 2.1   & 27.3  & 2.8   & 31.9  & 2.7   & 31.6  & 11.3  & 17.8  & 0.0   & 16.5  \\
        UGround-V1-7B~\cite{gounavigating2025} & 15.8  & 1.2   & 51.9  & 2.8   & 47.5  & 9.7   & 57.6  & 14.5  & 60.5  & 13.2  & 38.3  & 7.9   & 31.1  \\
        UI-TARS-7B~\cite{qin2025ui} & 20.8  & 9.4   & 58.4  & 12.4  & 50.0  & 9.1   & 63.9  & 31.8  & 63.3  & 20.8  & 30.8  & 16.9  & 35.7  \\
        TianXi-Action-7B~\cite{tang2025sea} & \textbf{76.0 } & \textbf{21.4 } & \textbf{61.6 } & \textbf{19.6 } & \textbf{45.2 } & \textbf{18.8 } & \textbf{80.6 } & \textbf{31.8 } & \textbf{84.2 } & \textbf{54.7 } & \textbf{57.9 } & \textbf{33.7 } & \textbf{51.9 } \\
        \midrule
        \multicolumn{14}{c}{\textbf{GUI-specific Models (RL)}} \\
        \midrule
        UI-R1-3B~\cite{lu2025ui} & 11.2  & 6.3   & 22.7  & 4.1   & 27.3  & 3.5   & 42.4  & 11.8  & 32.2  & 11.3  & 13.1  & 4.5   & 17.8  \\
        UI-R1-E-3B~\cite{lu2025ui} & 37.1  & 12.5  & 46.1  & 6.9   & 41.9  & 4.2   & 56.9  & 21.8  & 65.0  & 26.4  & 32.7  & 10.1  & 33.5  \\
        GUI-R1-7B~\cite{luo2025gui} & 23.9  & 6.3   & 49.4  & 4.8   & 38.9  & 8.4   & 55.6  & 11.8  & 58.7  & 26.4  & 42.1  & 16.9  & - \\
        InfiGUI-R1-3B~\cite{liu2025infigui} & 33.0  & 14.1  & 51.3  & 12.4  & 44.9  & 7.0   & 58.3  & 20.0  & 65.5  & 28.3  & 43.9  & 12.4  & 35.7  \\
        GUI-G1-3B~\cite{zhou2025gui} & 39.6  & 9.4   & 50.7  & 10.3  & 36.6  & 11.9  & 61.8  & 30.0  & 67.2  & 32.1  & 23.5  & 10.6  & 37.1  \\
        SE-GUI-7B~\cite{yuan2025enhancing} & 51.3  & 42.2  & 68.2  & 19.3  & 57.6  & 9.1   & 75.0  & 28.2  & 78.5  & 43.4  & 49.5  & 25.8  & 47.3  \\
        GUI-G2-7B~\cite{tang2025gui} & \textbf{55.8 } & \textbf{12.5 } & \textbf{68.8 } & \textbf{17.2 } & \textbf{57.1 } & \textbf{15.4 } & \textbf{77.1 } & \textbf{24.5 } & \textbf{74.0 } & \textbf{32.7 } & \textbf{57.9 } & \textbf{21.3 } & \textbf{47.5 } \\
        \midrule
        \multicolumn{14}{c}{\textbf{Test-Time Methods}} \\
        \midrule
        GUI-RC~\cite{du2025test} & -     & -     & -     & -     & -     & -     & -     & -     & -     & -     & -     & -     & 41.2  \\
        DiMo-GUI-7B~\cite{wu2025dimo} & 66.9  & 21.4  & 60.6  & 21.7  & 50.3  & 14.1  & 68.1  & 21.8  & 80.8  & 52.8  & 69.2  & 28.1  & 49.7  \\
        BAMI-7B & \textcolor[rgb]{ .753,  0,  0}{\textbf{81.8 }} & \textcolor[rgb]{ .753,  0,  0}{\textbf{26.9 }} & \textcolor[rgb]{ .753,  0,  0}{\textbf{68.2 }} & \textcolor[rgb]{ .753,  0,  0}{\textbf{23.8 }} & \textcolor[rgb]{ .753,  0,  0}{\textbf{58.4 }} & \textcolor[rgb]{ .753,  0,  0}{\textbf{29.7 }} & \textcolor[rgb]{ .753,  0,  0}{\textbf{77.8 }} & \textcolor[rgb]{ .753,  0,  0}{\textbf{36.4 }} & \textcolor[rgb]{ .753,  0,  0}{\textbf{83.6 }} & \textcolor[rgb]{ .753,  0,  0}{\textbf{60.4 }} & \textcolor[rgb]{ .753,  0,  0}{\textbf{72.9 }} & \textcolor[rgb]{ .753,  0,  0}{\textbf{33.3 }} & \textcolor[rgb]{ .753,  0,  0}{\textbf{57.8 }} \\
        \bottomrule
        \end{tabular}
    \label{tab:sota_sspro}
\end{table*}
\begin{table*}[t]
    \small
    \setlength{\tabcolsep}{3.5pt}
    \renewcommand{\arraystretch}{0.9}
    \centering
    \caption{Comparison with different baseline models on ScreenSpot-Pro.}
    \vspace{-2mm}
        \begin{tabular}{l|cc|cc|cc|cc|cc|cc|c}
        \toprule
        \multicolumn{1}{c|}{\multirow{2}[4]{*}{\textbf{Grounding Model}}} & \multicolumn{2}{c|}{\textcolor[rgb]{ .2,  .2,  .2}{\textbf{Development }}} & \multicolumn{2}{c|}{\textcolor[rgb]{ .2,  .2,  .2}{\textbf{Creative}}} & \multicolumn{2}{c|}{\textbf{CAD}} & \multicolumn{2}{c|}{\textcolor[rgb]{ .2,  .2,  .2}{\textbf{Scientific}}} & \multicolumn{2}{c|}{\textcolor[rgb]{ .2,  .2,  .2}{\textbf{Office}}} & \multicolumn{2}{c|}{\textcolor[rgb]{ .2,  .2,  .2}{\textbf{OS}}} & \multirow{2}[4]{*}{\textbf{Avg.}} \\
    \cmidrule{2-13}          & \textbf{Text} & \textbf{Icon} & \textbf{Text} & \textbf{Icon} & \textbf{Text} & \textbf{Icon} & \textbf{Text} & \textbf{Icon} & \textbf{Text} & \textbf{Icon} & \textbf{Text} & \textbf{Icon} &  \\
        \midrule
        UGround-7B~\cite{gounavigating2025} & 14.2  & 1.6   & 26.6  & 2.1   & 27.3  & 2.8   & 31.9  & 2.7   & 31.6  & 11.3  & 17.8  & 0.0   & 16.5  \\
        \textbf{ + BAMI} & \textbf{48.7 } & \textbf{5.5 } & \textbf{46.5 } & \textbf{7.7 } & \textbf{18.3 } & \textbf{4.7 } & \textbf{54.9 } & \textbf{14.6 } & \textbf{52.5 } & \textbf{18.9 } & \textbf{42.8 } & \textbf{9.4 } & \textbf{30.0 } \\
        \midrule
        OS-Atlas-7B~\cite{wu2024atlas} & 12.2  & 4.7   & 33.1  & 1.4   & 28.8  & 2.8   & 37.5  & 7.3   & 33.9  & 5.7   & 27.1  & 4.5   & 18.9  \\
        \textbf{ + BAMI} & \textbf{66.2 } & \textbf{16.6 } & \textbf{58.6 } & \textbf{16.1 } & \textbf{36.0 } & \textbf{10.9 } & \textbf{55.6 } & \textbf{17.3 } & \textbf{68.4 } & \textbf{22.6 } & \textbf{56.8 } & \textbf{17.1 } & \textbf{41.6 } \\
        \midrule
        UI-TARS-1.5-7B~\cite{qin2025ui} & 50.0  & 14.5  & 56.6  & 13.3  & 37.6  & 12.5  & 66.0  & 22.7  & 76.3  & 34.0  & 55.6  & 16.9  & 40.8  \\
        \textbf{ + BAMI} & \textbf{71.4 } & \textbf{22.1 } & \textbf{68.2 } & \textbf{21.7 } & \textbf{49.8 } & \textbf{14.1 } & \textbf{77.8 } & \textbf{23.6 } & \textbf{82.5 } & \textbf{41.5 } & \textbf{69.1 } & \textbf{24.2 } & \textbf{51.9 } \\
        \midrule
        TianXi-Action-7B~\cite{tang2025sea} & 76.0  & 21.4  & 61.6  & 19.6  & 45.2  & 18.8  & 80.6  & 31.8  & 84.2  & 54.7  & 57.9  & 33.7  & 51.9  \\
        \textbf{ + BAMI (GPT-5)} & \textbf{81.8 } & \textbf{26.9 } & \textbf{68.2 } & \textbf{23.8 } & \textbf{58.4 } & \textbf{29.7 } & \textbf{77.8 } & \textbf{36.4 } & \textbf{83.6 } & \textbf{60.4 } & \textbf{72.9 } & \textbf{33.3 } & \textbf{57.8 } \\
        \textbf{ + BAMI (Local)} & \textbf{80.5 } & \textbf{26.9 } & \textbf{66.7 } & \textbf{21.0 } & \textbf{53.8 } & \textbf{28.1 } & \textbf{78.5 } & \textbf{34.6 } & \textbf{83.1 } & \textbf{62.3 } & \textbf{71.3 } & \textbf{31.6 } & \textbf{56.2 } \\
        \bottomrule
        \end{tabular}
    \label{tab:diff-basemodel}
\end{table*}

\begin{figure*}[t]
    \centering
    \begin{subfigure}{0.51\textwidth}
        \centering
        \includegraphics[width=\textwidth]{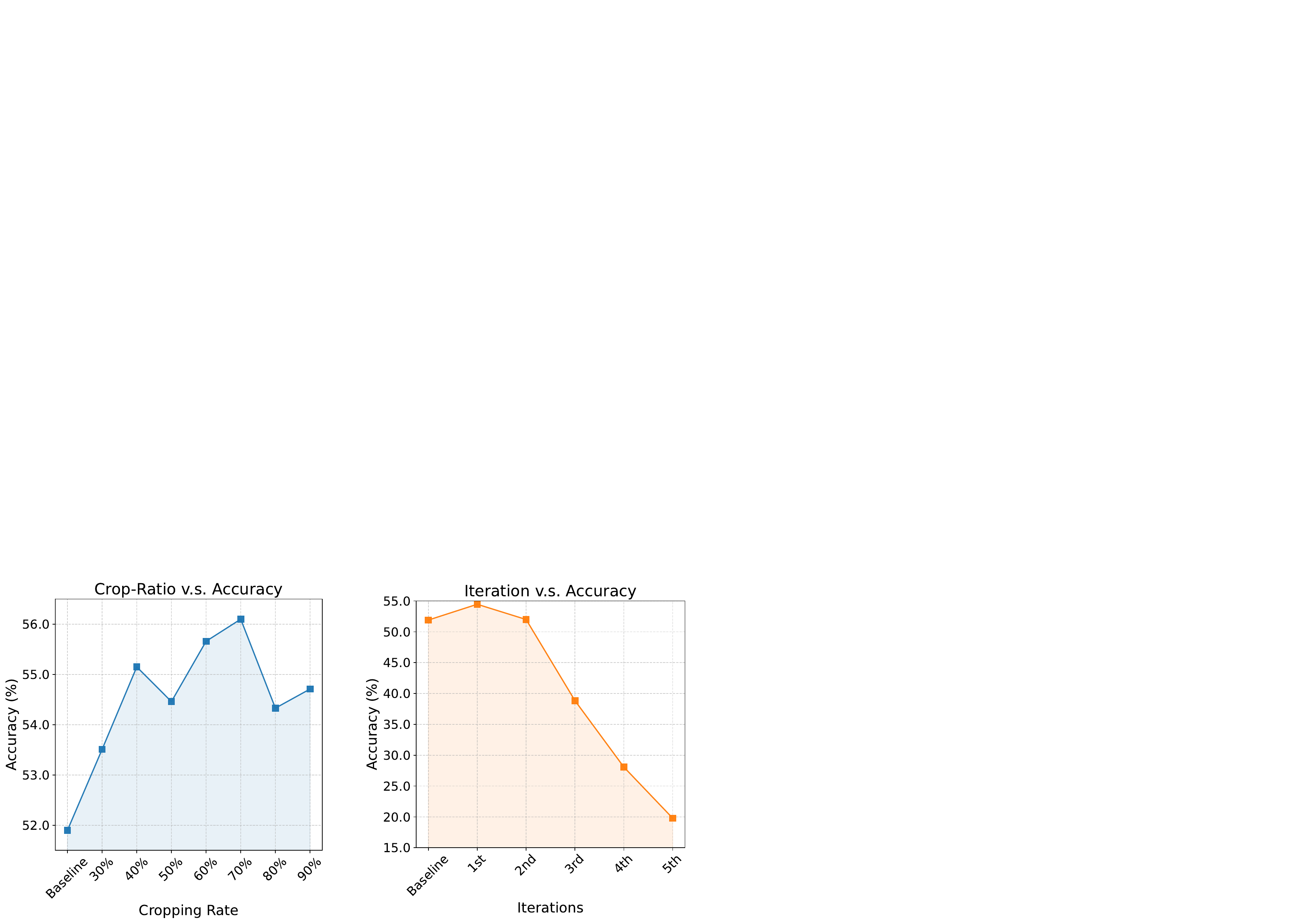}
        \caption{Ablations on crop ratio and iteration number.}
        \label{fig:prec_a}
    \end{subfigure}
    \hfill
    \begin{subfigure}{0.48\textwidth}
        \centering
        \includegraphics[width=\textwidth]{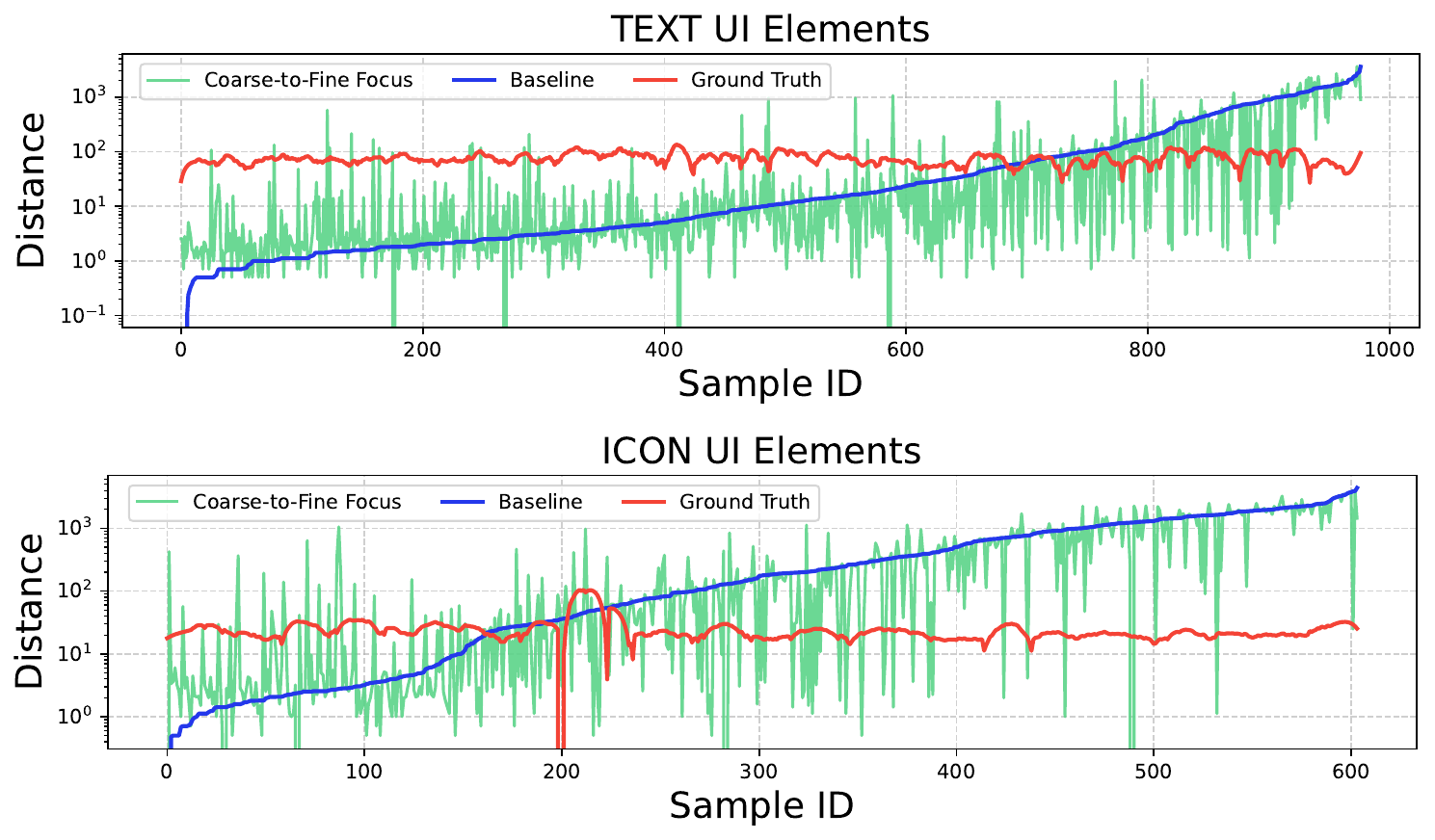}
        \caption{Impact of different target types.}
        \label{fig:prec_b}
    \end{subfigure}
    \vspace{-2mm}
    \caption{\textbf{Ablations on accuracy bias elimination.} (a) Effect of crop ratio and iteration count. (b) Performance across different target types.}
    \label{fig:prec_ablation}
    \vspace{-4mm}
\end{figure*}

\subsection{Comparison with SOTA}
We first evaluated state-of-the-art grounding methods on the complex ScreenSpot-Pro dataset, with results summarized in Table~\ref{tab:sota_sspro}. 
All models were categorized into three groups based on their training or deployment paradigms: supervised fine-tuning (SFT) training, reinforcement learning (RL) training, and test-time inference. 
Among 7B-scale models, our BAMI method achieved the best performance on the ScreenSpot-Pro dataset, attaining an accuracy of 57.8\%. 
Built upon TianXi-Action-7B~\cite{tang2025sea}, our model delivers a 5.9\% accuracy improvement. 
We present the consistent improvements of BAMI across different base models in Table~\ref{tab:diff-basemodel}. 
Additionally, we conducted experiments on the ScreenSpot-V2 dataset, with detailed results in the supplementary material, which demonstrates that our method outperforms all baseline models to varying extents. 
Furthermore, Table~\ref{tab:abl_combined} (left) dissects two key manipulations of BAMI, where ``PB'' and ``AB'' denote precision bias and ambiguity bias elimination, respectively. 
It can be observed that both manipulations independently yield significant improvements in model accuracy.
\subsection{Ablation Studies}
This section presents a series of ablation experiments to validate the effectiveness of the accuracy bias and ambiguity bias elimination manipulations in BAMI 
and explores the impact of different parameter settings on model performance.
\subsubsection{Accuracy Bias Elimination}
\paragraph{Impact of Crop Ratio and Iteration Number} 
We first investigate the effects of the number of crops and the crop ratio $\lambda$ on the elimination of accuracy bias, as illustrated in Figure~\ref{fig:prec_a}.
For the crop ratio, the number of iterations is fixed at 2. 
When the crop ratio exceeds 40\%, the elimination effect on precision bias becomes relatively significant. 
If the crop ratio is set too aggressively (i.e., less than 40\%), it may lead to the cropping of crucial contextual information, compromising model performance.
For the number of iterations, the crop ratio is fixed at 50\%. 
It can be observed that 2 iterations are sufficient to eliminate precision bias. 
Excessive iterations may result in an overly large overall cropping ratio, which conversely degrades the performance.
\paragraph{Impact of Target Type} 
We also investigate whether BAMI exhibits selectivity across different targets, as illustrated in Figure~\ref{fig:prec_b}.
We calculated the Euclidean distance between predictions and ground truth,
where the blue line represents the baseline and the green line represents BAMI.
The red line denotes the distance between corner points and the center of the ground truth bounding box.
If a prediction line lies below the red line, the prediction is highly likely correct.
We sorted the baseline results in ascending order for ease of observation.
For both text and icon types,
the bias distance of BAMI's predictions is mostly smaller than that of the baseline.
This indicates that BAMI has no selective preference for predicted targets and possesses universality.
\subsubsection{Ambiguity Bias Elimination}
\begin{table}[t]
  \centering
  \small
  \renewcommand{\arraystretch}{1.0}
  \setlength{\tabcolsep}{4pt}
  \caption{Ablation studies on BAMI components and prompt design.}
  \label{tab:abl_combined}
  \begin{tabular}{l|cc|c|l|cc|c}
    \toprule
    \multicolumn{4}{c|}{\textbf{Manipulations}} & \multicolumn{4}{c}{\textbf{Prompt Design}} \\
    \midrule
    \textbf{Setting} & \textbf{PB} & \textbf{AB} & \textbf{Acc.} & \textbf{Setting} & \textbf{CoT} & \textbf{KP} & \textbf{Acc.} \\
    \midrule
    Baseline &  &  & 51.9 & Baseline &  &  & 51.9 \\
    + C2F Focus & \checkmark &  & 55.2 & + Vanilla &  &  & 55.7 \\
    + Cand. Sel. &  & \checkmark & 54.3 & + w/ CoT  & \checkmark &  & 57.0 \\
    \textbf{+ BAMI} & \checkmark & \checkmark & \textbf{57.8} & \textbf{+ w/ CoT\&KP} & \checkmark & \checkmark & \textbf{57.8} \\
    \bottomrule
  \end{tabular}
  \vspace{-2mm}
\end{table}


\paragraph{Impact of Prompt Design} 
A key reason for ambiguity bias lies in the fact that MLLMs prioritize candidate outcomes based on edit distance. 
Therefore, it is crucial to inject priority priors in the Euclidean space through prompt design. 
We conducted ablation experiments on two important prompt structures in BAMI, as shown in Table~\ref{tab:abl_combined} (right). 
``CoT'' denotes the chain-of-thought-style prompt, which aims to enable the correction model to make selections in a more granular manner. 
``KP'' stands for key principle, a critical component for injecting coordinate space priority priors into the selection process. 
Experimental results demonstrate that injecting Euclidean space priors into the correction model significantly enhances the accuracy of BAMI.
\paragraph{Impact of Correction Model Selection} 
We investigated the impact of correction model selection, as presented in Table~\ref{tab:abl_correction}. 
Among online APIs, GPT-5 and Gemini-2.5-Pro achieved the best performance, enabling an overall accuracy of over 57\%. 
All tested models contributed to performance improvement, indicating BAMI's robustness across different correction models.
To address privacy requirements and enable independent deployment, we trained a local Qwen3-VL-8B correction model (~8B parameters, matching grounding models' scale) via LoRA fine-tuning on 128K dual-box samples. 
As shown in Table~\ref{tab:abl_correction} and Table~\ref{tab:diff-basemodel}, it achieves 56.2
Training details are in the supplementary material.

\begin{table}[t]
  \setlength{\tabcolsep}{4pt}
  \centering
  \caption{Impact of different correction models (online and offline).}
    \begin{tabular}{l|l|c}
    \toprule
    \textbf{Category} & \textbf{Correction Model} & \textbf{Accuracy} \\
    \midrule
    \multicolumn{2}{l|}{Baseline} & 51.9 \\
    \midrule
    \multirow{4}{*}{\textit{Online APIs}} 
    & Doubao-Seed-1.6-Flash & 55.3 \\
    & GLM-4.5V & 55.9 \\
    & Qwen-VL-Max & 56.4 \\
    & Gemini-2.5-Pro & 57.2 \\
    & \textbf{GPT-5} & \textbf{57.8} \\
    \midrule
    \textit{Local Model} & Qwen3-VL-8B (Ours) & 56.2 \\
    \bottomrule
    \end{tabular}
  \label{tab:abl_correction}
  \vspace{-2mm}
\end{table}
\section{Discussion}
This paper investigates how to improve GUI grounding in a training-free manner.
Through \textbf{MPD}, we identify that most errors on complex GUIs stem from inductive bias, mainly accuracy bias and ambiguity bias.
\textbf{BAMI} extends the model's reasoning process through structured inference with two critical manipulations, coarse-to-fine focus and candidate selection, which substantially alleviate these biases.
On ScreenSpot-Pro, BAMI improves TianXi-Action-7B from 51.9\% to 57.8\%, showing clear advantages over contemporary training-free methods in both effectiveness and efficiency.
To address privacy concerns and avoid relying on the extra knowledge implicitly introduced by external APIs, we further train a local correction-model variant based on Qwen3-VL-8B using public data, which reaches 56.2\%.
In particular, this local model uses a parameter scale comparable to the grounding models themselves, demonstrating that effective correction can be achieved without requiring significantly larger architectures.
Beyond this practical variant, we will further investigate how the model's inductive preferences diverge from real-world GUI scenarios, with the goal of developing more generalizable solutions.

\section*{Acknowledgement}

This work was supported in part by the Beijing Natural Science Foundation under Grant No. L247009, and the National Natural Science Foundation of China under Grant 62125603, Grant 62336004, Grant 62321005.
{
    \small
    \bibliographystyle{ieeenat_fullname}
    \bibliography{main}
}

\appendix
\onecolumn
\startcontents[sections]
\begin{center}
    \Large\textbf{\thetitle}\\
    \vspace{0.5em}
    \Large\textbf{Supplementary Material}\\
    \vspace{1em}
    \large\textbf{Table of Contents for Supplementary Material}
\end{center}
\vspace{0.5em}
\rule{\linewidth}{0.5pt}
\vspace{-0.5em}
\printcontents[sections]{l}{1}{
    \setcounter{tocdepth}{2}
}
\rule{\linewidth}{0.5pt}
\vspace{1em}

\section{Usage of Large Models in Paper Writing} \label{sec:llm_usage}

During the conduct of this research, we utilized the GPT-5 for auxiliary support, primarily encompassing the following two aspects:
\begin{itemize}
    \item \textbf{Manuscript Polishing:} Leveraging the text generation capability of GPT-5, we polished the draft of this manuscript, focusing on correcting grammatical errors, addressing expression inconsistencies, and other related issues.
    It should be emphasized that all content of the manuscript was still manually composed; the LLM was not involved in formulating the research logic of the paper.
    Additionally, all text generated by the LLM underwent manual review and revision to ensure its quality and accuracy.
    \item \textbf{Literature Survey:} We employed the knowledge retrieval capability (Retrieval-Augmented Generation, RAG) of GPT-5 to search for relevant literature.
    To guarantee retrieval accuracy, all retrieved literature was subject to manual review and verification.
    Subsequently, we screened out literature relevant to the research topic, followed by thorough reading and systematic organization of the selected materials.
\end{itemize}

\section{Details of the Proposed Methods}

\subsection{Detailed Algorithm of MPD Attribution}

To investigate the root causes of errors in grounding models,
we propose a method for rapidly computing the decision attribution of models, namely \textbf{Masked Prediction Distribution (MPD) Attribution}. The detailed steps of this algorithm are presented as follows:

\begin{algorithm}[t]
\caption{Masked Prediction Distribution (MPD) Attribution Algorithm}
\label{alg:mpd}
\begin{algorithmic}[1]
    \REQUIRE GUI image $I$, query $q$, grid size $(M,N)$, number of samples $K$
    \ENSURE Set of predicted points $\mathcal{P} = \{(x_c^{(k)}, y_c^{(k)})\}_{k=1}^K$
    \STATE Partition the image $I$ into $M \times N$ grid blocks $\{B_{i,j}\}_{i=1,j=1}^{M,N}$
    \FOR{$k = 1$ to $K$}
        \STATE Randomly select a masking ratio $\alpha$ and sample $\lfloor \alpha \cdot M \cdot N \rfloor$ grid blocks to mask
        \STATE Generate the masked image $I^{(k)}$, where masked regions are filled with zero vectors
        \STATE Compute the model prediction: $t^{(k)} = f(q, I^{(k)})$
        \STATE Extract the center coordinates: $(x_c^{(k)}, y_c^{(k)})$
    \ENDFOR
    \STATE Visualize all predicted points $\{(x_c^{(k)}, y_c^{(k)})\}_{k=1}^K$ as a scatter plot
\end{algorithmic}
\end{algorithm}

\section{Experimental Details}

\subsection{Prompt Design} \label{sec:app_prompt_design}

The design of prompts is crucial for injecting prior information of coordinate space into the candidate box selection process.
In the experiments presented in Table 4 (main paper), we compare prompts with different content.
Among these, the vanilla prompt is as follows:

\begin{figure*}[t]
\centering
\begin{minipage}{\textwidth}
\begin{lstlisting}[title=Prompt]
You are comparing two images to determine which one better fulfills the user's intent.

User Command: "{user_query}"

Image 1: Shows a GUI element marked with a green box labeled "1"
Image 2: Shows a GUI element marked with a red box labeled "2"

Your task: Determine which image shows the element that will best fulfill the user's command.

**OUTPUT FORMAT**:
<answer>1 or 2</answer>"""
\end{lstlisting}
\end{minipage}
\end{figure*}

This simplistic prompt design fails to rectify the model's ambiguity bias.
Therefore, in our BAMI method, we incorporate two critical structures—chain of thought and key principle—to enhance the model's understanding of prior information regarding the coordinate space.
The final prompt we employed is presented as follows:

\begin{figure*}[t]
\centering
\begin{minipage}{\textwidth}
\begin{lstlisting}[title=Prompt]
You are comparing two images to determine which one better fulfills the user's intent.

User Command: "{user_query}"

Image 1: Shows a GUI element marked with a green box labeled "1"
Image 2: Shows a GUI element marked with a red box labeled "2"

Your task: Determine which image shows the element that will best fulfill the user's command.

ANALYSIS APPROACH:
1. Examine what GUI element is highlighted in each image
2. Consider which element better matches the user's intent
3. Think about standard GUI patterns and user expectations
4. Choose the image that shows the more appropriate interaction target

KEY PRINCIPLES:
- Focus on the functional purpose of the highlighted elements
- Consider standard UI patterns (buttons for actions, text fields for input, etc.)
- Choose interactive elements over static text/labels
- If one shows a selected state and the other shows normal state, prefer the normal state
- ELEMENT QUALITY HIERARCHY (best to worst):
   - Icon + Text together (most informative and complete)
   - Complete icon alone (clear visual indicator)
   - Complete text alone (readable label)
   - Multiple elements in one box OR incomplete elements (ambiguous target)

COMMON PITFALLS TO AVOID:
    - Don't choose based on keyword matching alone
    - Don't overlook the user's actual goal in favor of literal interpretation

Remember: Provide SPECIFIC analysis based on what you actually observe, not generic descriptions.

**OUTPUT FORMAT**:
<analysis>
Image 1: [Describe what element is highlighted and its purpose]
Image 2: [Describe what element is highlighted and its purpose]
Comparison: [Explain which better serves the user's intent and why]
</analysis>

<answer>1 or 2</answer>
<reason>Brief explanation of why this image shows the better choice</reason>
\end{lstlisting}
\end{minipage}
\end{figure*}

\subsection{Model Inference Details}

The models employed in this study can be broadly categorized into two types:
\begin{itemize}
    \item \textbf{Bounding box-output models}: Such as OS-Atlas-7B~\cite{wu2024atlas} and TianXi-Action-7B~\cite{tang2025sea}
    \item \textbf{Click point-output models}: Such as UGround~\cite{gounavigating2025} and UI-TARS-1.5-7B~\cite{qin2025ui}
\end{itemize}
For \textbf{bounding box-output models}, the implementation of masked prediction is straightforward, only the pixels within the output bounding boxes need to be masked.
In contrast, for \textbf{click point-output models}, we first expand the region around each click point by a fixed number of pixels (e.g., 25 pixels) in the up, down, left, and right directions, and then mask the expanded region.

\subsection{Local Correction Model Training}

To enable offline deployment of BAMI, we trained a specialized correction model based on Qwen3-VL-8B using LoRA fine-tuning.
The training dataset contains 128,487 dual-box samples automatically generated via our five-step pipeline, sourced from GUIAct (70K samples) and Desktop domain (423K samples, then downsampled).
Labels are determined by comparing against ground truth using dual criteria: IoU $\geq$ 0.5 or center point within GT bbox.
When both boxes satisfy the criteria, we prioritize bbox1 (baseline) to reflect regrounding's role as a fallback mechanism, resulting in a 92:8 label distribution (bbox1:bbox2).

We fine-tune only the language model component via LoRA (rank $r$=128, alpha $\alpha$=256, dropout=0.05) while freezing the vision encoder and projection layers, yielding approximately 200M trainable parameters (2.5\% of total).
This design leverages the pre-trained visual understanding while adapting the decision-making capability for dual-box selection.
Training employs 8$\times$ A100 80GB GPUs with DeepSpeed ZeRO-2 optimization, effective batch size 128, learning rate 1e-4 with cosine annealing, for 3 epochs (approximately 12 hours).
The model uses the same 24-line instruction prompt as GPT-5 (detailed in Section~\ref{sec:app_prompt_design}) to ensure consistent task understanding and incorporate GUI-specific priors.

On ScreenSpot-Pro evaluation, the local model selects bbox2 in 9.7\% of cases, closely matching the training distribution (8\%), indicating proper learning of the selection strategy without overfitting to always choose baseline.
The 56.2\% accuracy demonstrates that comparable-scale models (~8B parameters) can effectively perform correction tasks without requiring significantly larger architectures.

\section{More Experiments}

\subsection{Comparison on ScreenSpot-V2}

In addition to validating the BAMI method on the ScreenSpot-Pro~\cite{li2025screenspot} dataset,
we also conducted validation on the simpler ScreenSpot-V2~\cite{wu2024atlas} dataset.
Unlike ScreenSpot-Pro, most grounding models already achieve satisfactory accuracy on ScreenSpot-V2;
this is attributed to the lower resolution of samples and the simpler elements contained in individual samples within the latter dataset.
When we applied the BAMI method to the OS-Atlas-7B and UI-TARS-1.5-7B models,
further performance improvements were observed.
However, the magnitude of these improvements is smaller than that achieved on the ScreenSpot-Pro dataset.

\begin{table*}[t]
\centering
\small
\caption{Comparison with various models on ScreenSpot-V2.}
\label{tab:sota_ssv2}
\begin{tabular}{l|cc|cc|cc|c}
    \toprule
    \multicolumn{1}{c|}{\multirow{2}[2]{*}{\textbf{Grounding Model}}} & \multicolumn{2}{c|}{\textbf{Mobile}} & \multicolumn{2}{c|}{\textbf{Desktop}} & \multicolumn{2}{c|}{\textbf{Web}} & \multirow{2}[2]{*}{\textbf{Avg.}} \\
          & \textbf{Text} & \textbf{Icon} & \textbf{Text} & \textbf{Icon} & \textbf{Text} & \textbf{Icon} &  \\
    \midrule
    InternVL-2-4B~\cite{chen2024internvl} & 9.2   & 4.8   & 4.6   & 4.3   & 0.9   & 0.1   & 4.3  \\
    Qwen2-VL-7B~\cite{wang2024qwen2} & 61.3  & 39.3  & 52.0  & 45.0  & 33.0  & 21.8  & 42.9  \\
    CogAgent~\cite{hong2024cogagent} & 67.0  & 24.0  & 74.2  & 20.0  & 70.4  & 28.6  & 47.4  \\
    SeeClick~\cite{cheng2024seeclick} & 78.0  & 52.0  & 72.2  & 30.0  & 55.7  & 32.5  & 53.4  \\
    OS-Atlas-4B~\cite{wu2024atlas} & 85.7  & 58.5  & 72.2  & 45.7  & 82.6  & 63.1  & 70.1  \\
    UGround-7B~\cite{gounavigating2025} & 82.8  & 82.8  & 82.8  & 63.6  & 80.4  & 70.4  & 73.3  \\
    \midrule
    OS-Atlas 7B~\cite{wu2024atlas} & 92.1  & 68.7  & 88.7  & 60.7  & 89.7  & 75.9  & 81.2 \\
    \textbf{ + BAMI} & \textbf{92.4} & \textbf{67.3} & \textbf{88.7} & \textbf{66.4} & \textbf{89.3} & \textbf{79.8} & \textbf{82.2} \\
    UI-TARS-1.5-7B~\cite{qin2025ui} & 94.1  & 80.6  & 88.7  & 76.4  & 88    & 84.2  & 86.4 \\
    \textbf{ + BAMI} & \textbf{94.1} & \textbf{80.6} & \textbf{88.7} & \textbf{76.4} & \textbf{88} & \textbf{84.7} & \textbf{86.5} \\
    \bottomrule
\end{tabular}
\end{table*}

\subsection{Why Masking Is Adopted Instead of Random Sampling?}

\begin{figure*}[t]
    \centering
    \begin{subfigure}{0.49\textwidth}
        \centering
        \includegraphics[width=\textwidth]{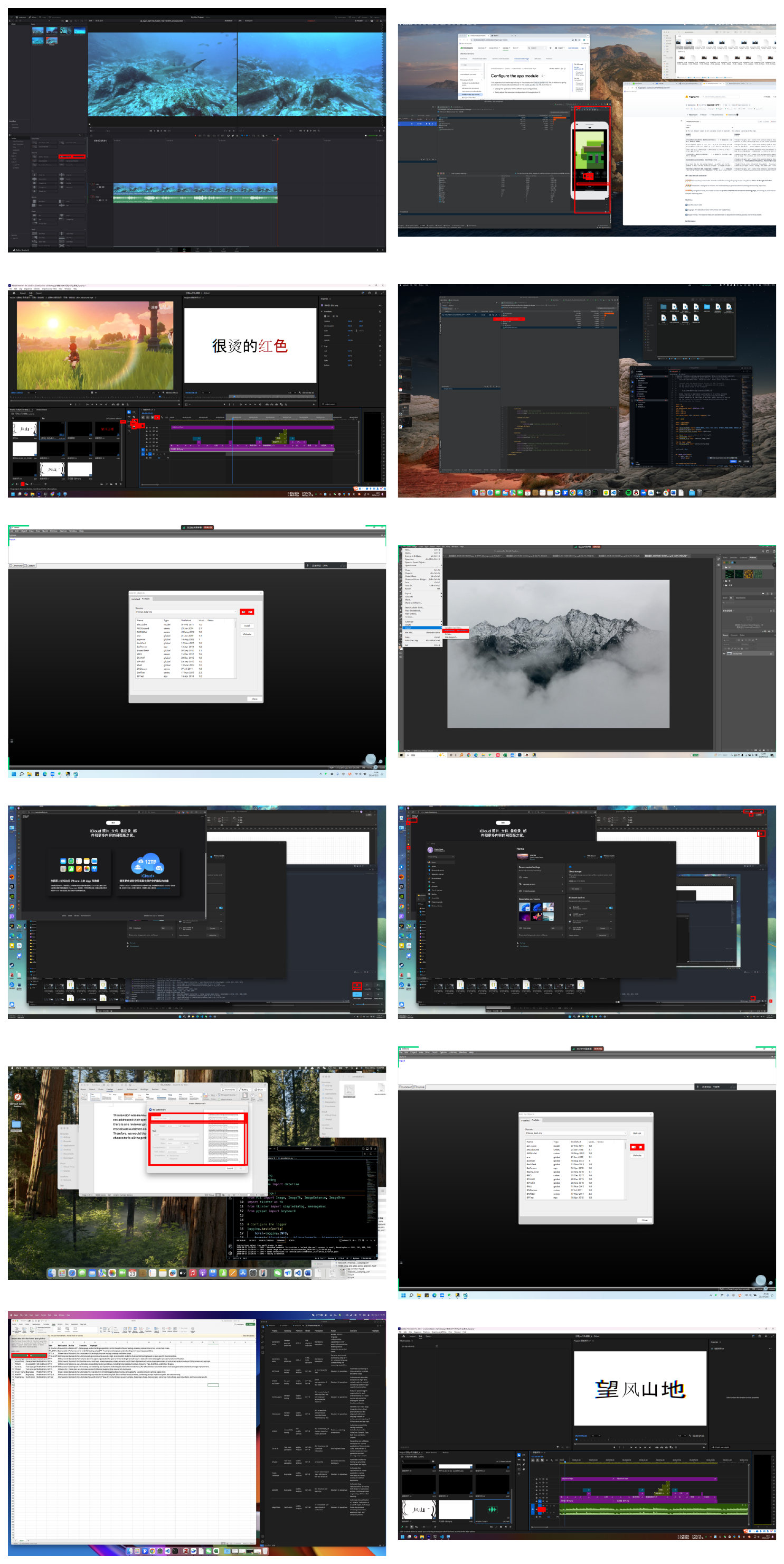}
        \caption{Candidate boxes with random sampling.}
        \label{fig:sample_random}
    \end{subfigure}
    \hfill
    \begin{subfigure}{0.49\textwidth}
        \centering
        \includegraphics[width=\textwidth]{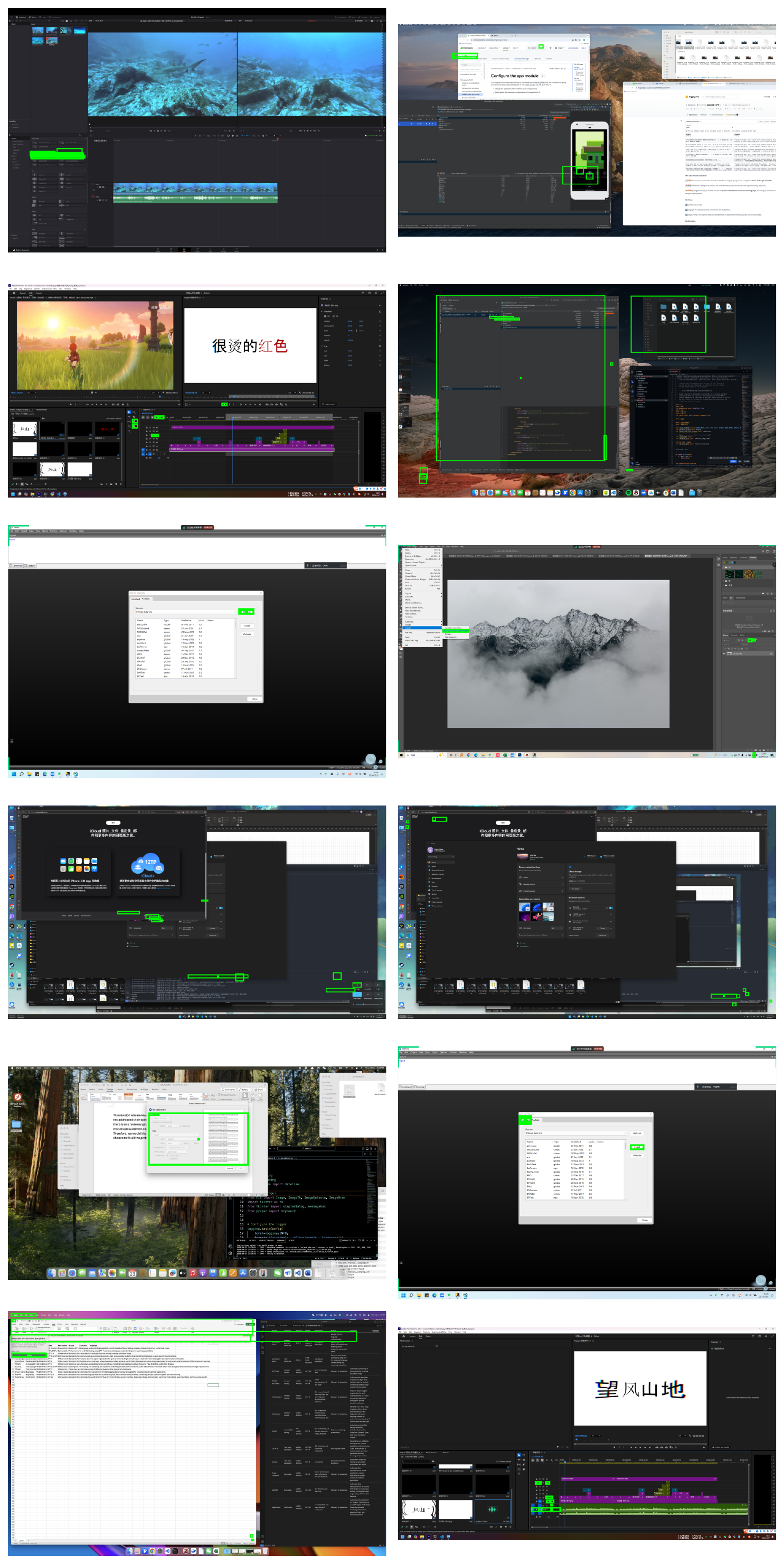}
        \caption{Candidate boxes with masked prediction.}
        \label{fig:sample_mask}
    \end{subfigure}
    \caption{Comparison of candidate box generation strategies.}
\end{figure*}

In conventional approaches for generating candidate detection boxes, random sampling is typically employed.
Specifically, when predicting the next token, instead of using the \texttt{torch.argmax} function to greedily select the token corresponding to the highest score,
top-k/top-p sampling methods are utilized to obtain candidate tokens.
However, our experiments reveal that in GUI grounding models during candidate box generation, the score difference between the top-1 token and top-2 token is substantial.
This directly leads to a significant issue: candidate boxes generated via random sampling tend to cluster in a single region.
As illustrated in Figure~\ref{fig:sample_random}, the red boxes represent candidate boxes obtained through random sampling. It is evident that these boxes exhibit almost complete overlap and lack diversity, which renders the subsequent selection process largely meaningless.

To address this limitation, we propose a masking strategy: pixels within the already predicted candidate boxes are masked first.
This ensures that subsequently predicted candidate boxes are mutually exclusive with the already predicted ones.
As shown in Figure~\ref{fig:sample_mask}, the green boxes are candidate boxes generated using the masked prediction method.
These boxes demonstrate significantly greater distribution diversity, thereby enhancing the upper performance limit of selection manipulation.

\subsection{More Visualizations of BAMI}

\begin{figure*}[t]
    \centering
    \includegraphics[width=0.9\textwidth]{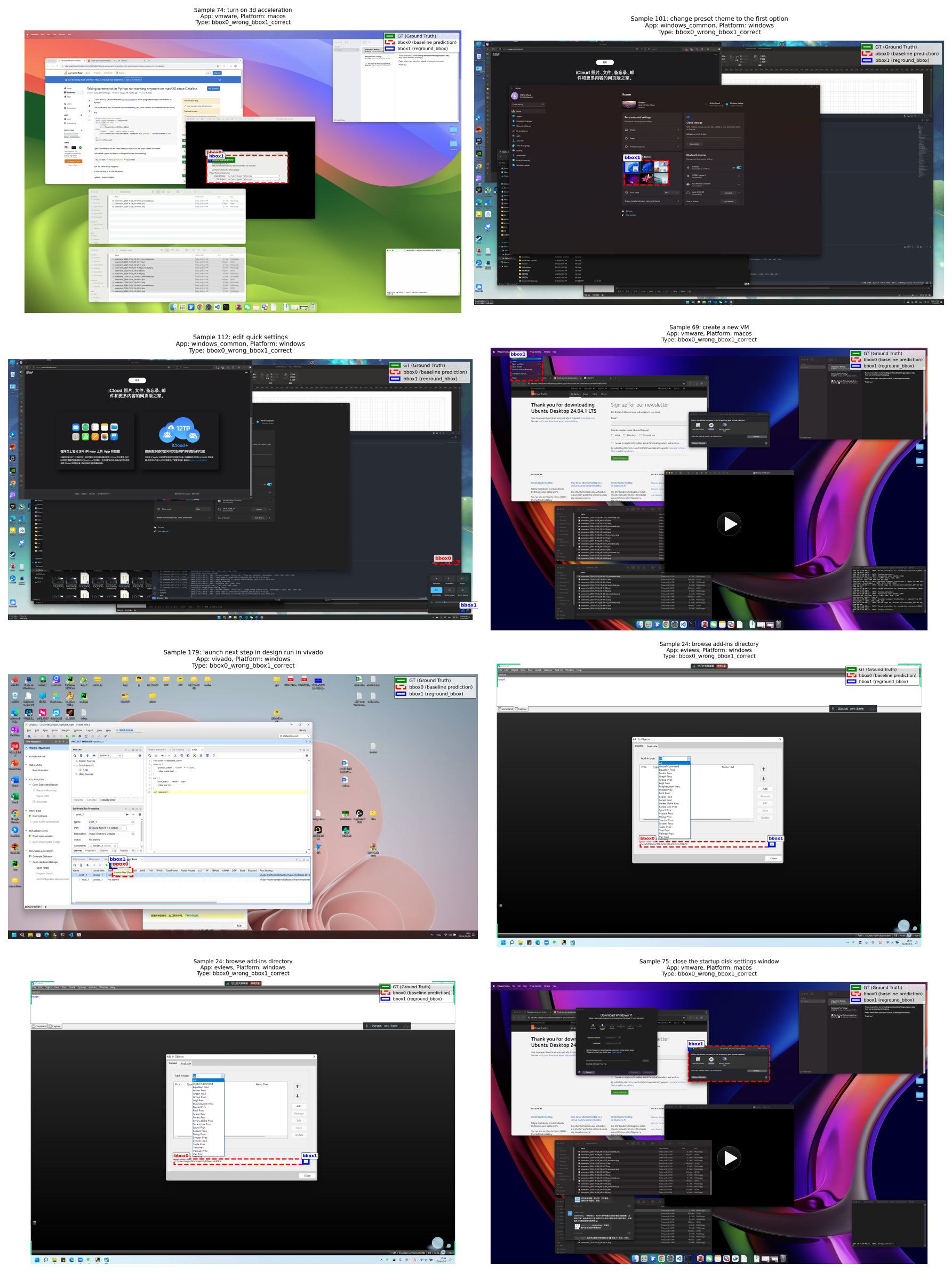}
    \caption{Visualizations of BAMI corrections.}
    \label{fig:app_vis}
\end{figure*}

To better demonstrate the process by which BAMI corrects the baseline model, 8 samples were randomly selected from cases where the baseline model made incorrect predictions but BAMI achieved accurate corrections, as shown in Figure~\ref{fig:app_vis}.
In the figure, green boxes represent ground truth, red boxes denote the baseline model's prediction results (incorrect), and blue boxes indicate the corrected results by BAMI (correct).
Specifically, BAMI utilized 2 candidate boxes in each prediction round of this experiment, with GPT-5 as the correction model.
In these samples, it can be observed that accurately predicting bounding boxes in accordance with user instructions is considerably challenging, as the figures contain substantial interfering information.
By alleviating precision bias and ambiguity bias, BAMI successfully achieves correct predictions in these samples.

\subsection{More Attribution Results}

\begin{figure*}[t]
    \centering
    \includegraphics[width=0.9\textwidth]{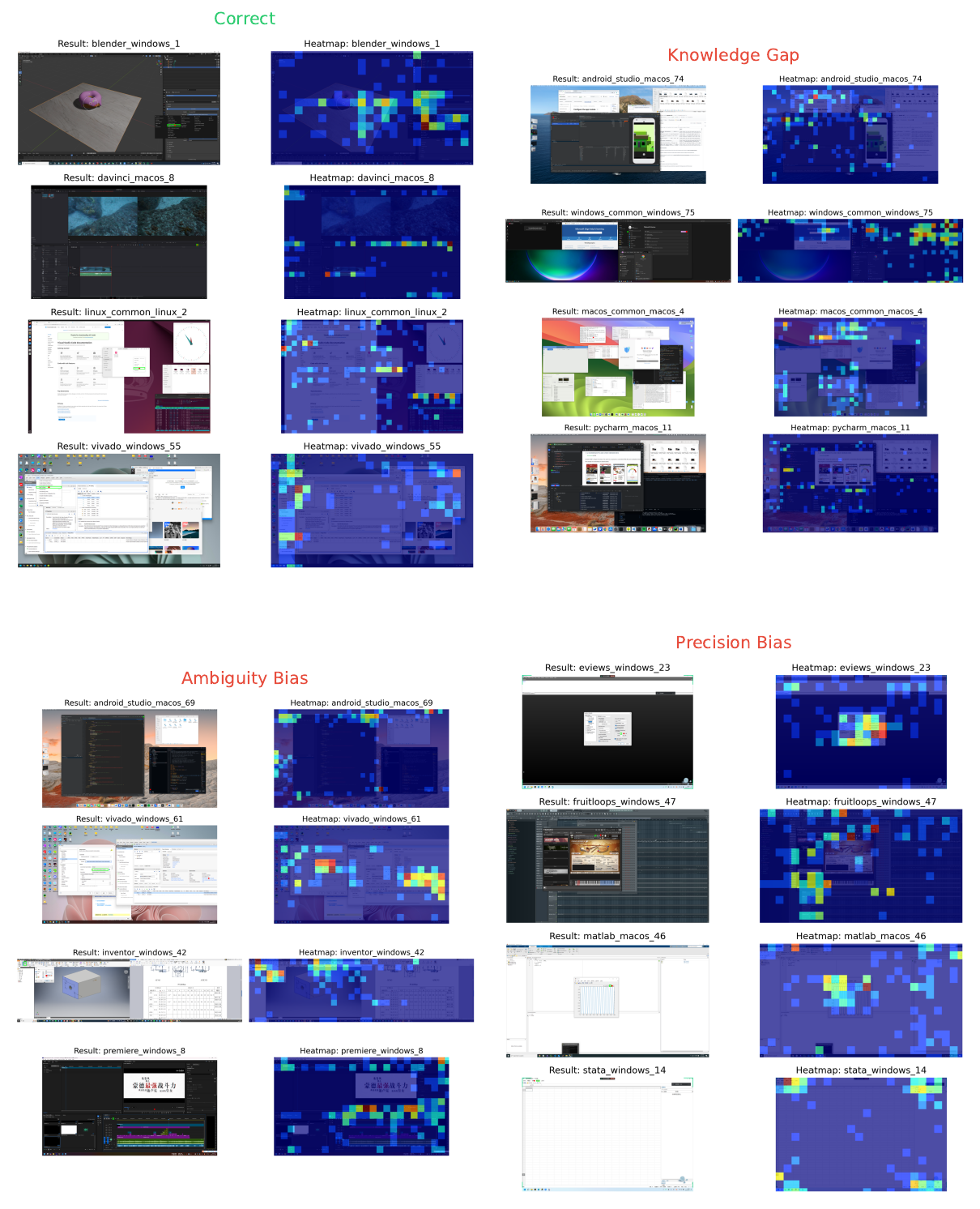}
    \caption{More Attributions Visualizations.}
    \label{fig:app_attr}
\end{figure*}

We present additional attribution results herein to comprehensively demonstrate the attribution capability of the Masked Prediction Distribution (MPD) method.
Specifically, we randomly selected samples from four categories (\texttt{Correct / Knowledge Gap / Precision Bias / Ambiguity Bias}) as illustrated in Figure~\ref{fig:app_attr}.

\end{document}